\documentclass[10pt,twocolumn,letterpaper]{article}

\usepackage[pagenumbers]{iccv} 

\definecolor{iccvblue}{rgb}{0.21,0.49,0.74}
\usepackage[pagebackref,breaklinks,colorlinks,allcolors=iccvblue]{hyperref}
\usepackage{amsmath} 
\usepackage{amssymb} 
\usepackage{graphicx} 
\usepackage{multirow} 
\usepackage{array}  
\usepackage{balance}
\usepackage{bm}

\usepackage{booktabs} 
\usepackage{soul} 
\usepackage{xcolor} 
\usepackage{colortbl}
\usepackage{utfsym}
\usepackage{pifont}
\hypersetup{
bookmarksnumbered=true,
}
\usepackage{bookmark}
\usepackage{fontawesome}
\sethlcolor{orange!30}
\definecolor{iccvblue}{rgb}{0.21,0.49,0.74}
\definecolor{red}{RGB}{192,0,0}
\definecolor{green}{RGB}{0,176,80}
\definecolor{blue}{RGB}{0,112,192}
\usepackage{makecell}
\let\bs\boldsymbol
\newcommand{\acronym}{\textsc{Garf}}
\newcommand{\dataset}{\textsc{Fractura}}
\usepackage{tikz}
\newcommand{\bcircled}[1]{%
\tikz[baseline=(char.base)]{ \node[shape=circle,draw,inner sep=1pt] (char)
{\textbf{#1}};%
}%
}

\def\paperID{11316} 
\def\confName{ICCV}
\def\confYear{2025}

\title{\textsc{Garf}: Learning Generalizable 3D Reassembly for Real-World Fractures}

\author{Sihang Li$^{1,}\footnotemark[1]$, Zeyu Jiang$^{1,}\footnotemark[1]$, Grace Chen$^{1,}\footnotemark[2]$, Chenyang Xu$^{1,}\footnotemark[2]$, Siqi Tan$^{1}$,  Xue Wang$^{1}$, Irving Fang$^{1}$,\\ Kristof Zyskowski$^{2}$, Shannon P. McPherron$^{3}$, Radu Iovita$^{1}$, Chen Feng\textsuperscript{1,\ding{41}}, Jing Zhang\textsuperscript{1,\ding{41}} \\
\\
\textsuperscript{1}New York University \quad \textsuperscript{2}Yale University \quad \textsuperscript{3}Max Planck Institute
}

\begin{document}
\twocolumn[{%
    \maketitle
    \vspace{-15pt}
    \centering
    \includegraphics[width=\textwidth]{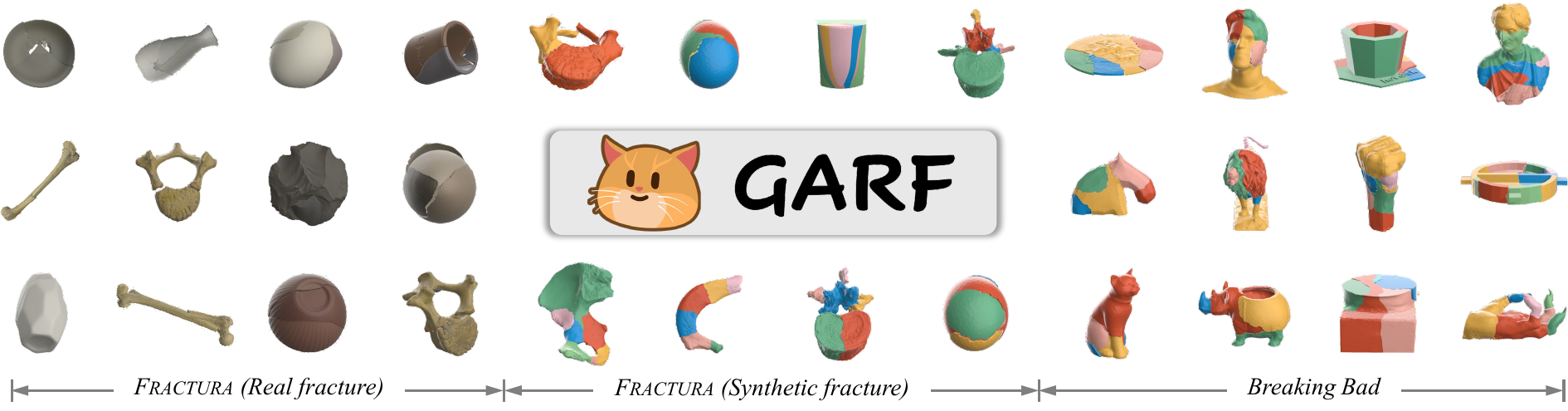}
    \vspace{-20pt}
    \captionof{figure}{We curate \dataset{}, a unique dataset presenting real-world fracture assembly challenges across scientific domains, including ceramics, bones, eggshells, and lithics. To tackle these challenges, we introduce \acronym{}, a \textit{generalizable} 3D reassembly framework designed to handle varying \textit{object shapes}, diverse \textit{fracture types}, and the presence of \textit{missing} or \textit{extraneous} fragments.}
    \label{fig:teaser}
    \vspace{10pt}
}]

{
  \renewcommand{\thefootnote}%
    {\fnsymbol{footnote}}
  \footnotetext[1]{, † Equal contribution.}
  \footnotetext{\thanks{\ding{41} Corresponding authors {\tt\small \{\href{mailto:z.jing@nyu.edu}{z.jing}, \href{mailto:cfeng@nyu.edu}{cfeng}\}@nyu.edu}.}
  
}

\begin{abstract}

\vspace{-6pt}
3D reassembly is a challenging spatial intelligence task with broad applications across scientific domains. While large-scale synthetic datasets have fueled promising learning-based approaches, their generalizability to different domains is limited. Critically, it remains uncertain whether models trained on synthetic datasets can generalize to real-world fractures where breakage patterns are more complex. To bridge this gap, we propose \acronym{}, a \textbf{g}eneralizable 3D re\textbf{a}ssembly framework for \textbf{r}eal-world \textbf{f}ractures. \acronym{} leverages fracture-aware pretraining to learn fracture features from individual fragments, with flow matching enabling precise 6-DoF alignments. At inference time, we introduce two-session
flow matching, improving robustness to unseen objects and varying numbers of fractures. In collaboration with archaeologists, paleoanthropologists, and ornithologists, we curate \dataset{}, a diverse dataset for vision and learning communities, featuring real-world fracture types across ceramics, bones, eggshells, and lithics. Comprehensive experiments have shown our approach consistently outperforms state-of-the-art methods on both synthetic and real-world datasets, achieving 82.87\% lower rotation error and 25.15\% higher part accuracy on the Breaking Bad Everyday dataset. This sheds light on training on synthetic data to advance real-world 3D puzzle solving, demonstrating its strong generalization across unseen object shapes and diverse fracture types. \acronym{}'s code, data and demo are available at \url{https://ai4ce.github.io/GARF/}.
\end{abstract}
    
\vspace{-15pt}
\section{Introduction}
\label{sec:intro}

We have long been captivated by questions of human origins~\cite{wilson2012social}: \textit{What are we? Where do we come from?} The answers to these fundamental questions lie in archaeological materials such as bones~\cite{callaway2017oldest}, ceramics~\cite{wang2019emergence}, and lithics~\cite{mcpherron2010evidence}. However, these artifacts are often highly fragmented and incomplete~\cite{callaway2017oldest, laughlin_149_2005}. Reassembling them requires placing each fragment in its correct \textit{position} and \textit{orientation} to restore a complete or functional entity~\cite{scarpellini2024diffassemble}. This process is not only time-consuming but also challenges the limits of human \textit{spatial intelligence}~\cite{scarpellini2024diffassemble, gardner2011frames, tsesmelis2025re}. For instance, an experimental study on lithic refitting reported a success rate of only 30\%, with experts performing only marginally better than novices~\cite{laughlin_experimental_2010}. Consequently, museum storerooms around the world remain filled with thousands of unassembled fragments, waiting to be pieced back together~\cite{tsesmelis2025re}.

Recently, the emergence of the large-scale dataset Breaking Bad~\cite{sellan2022breaking} has brought new hope to this domain, fueling the development of data-intensive reassembly methods~\cite{wu2023leveraging, villegas2023matchmakernet, zhang2024scalable, li2024geometric, lu2024jigsaw, cui2024phformer, scarpellini2024diffassemble, wang2024puzzlefusion++, lu2024jigsaw++, xu2024fragmentdiff, lee20243d, kim2024fracture}. While PuzzleFusion++ achieves state-of-the-art (SOTA) performance on the \textit{everyday} subset, its generalization degrades significantly on the \textit{artifact} subset due to its reliance on global geometry learning~\cite{wang2024puzzlefusion++}. More critically, the Breaking Bad dataset is generated via physics-based simulation~\cite{sellan2023breaking}, raising a fundamental question: \textit{can models trained on synthetic breakage patterns generalize to real-world fractures}?

\begin{figure*}[t]
    \centering
    \includegraphics[width=\linewidth]{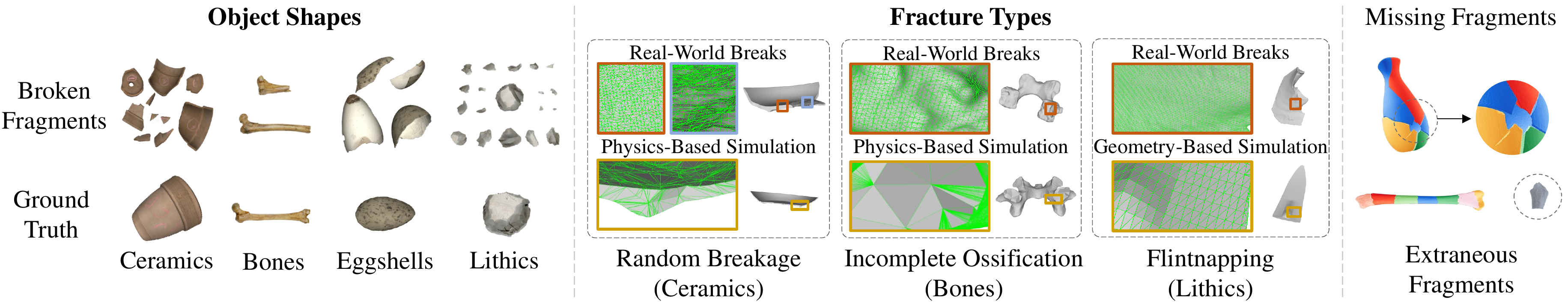}
    \vspace{-20pt}
    \caption{\textbf{Characteristics and Challenges of \dataset{}.} (i) Diverse fracture types including two synthetic and three real-world types, across ceramics, bones, eggshells, and lithics. (ii) Real-world challenges such as missing or extraneous fragments.}
    \vspace{-14pt}
    \label{fig: Dataset}
\end{figure*}

To answer this question, we identify two major real-world fracture challenges and curate \textsc{Fractura}, a dataset capturing key complexities: (\textbf{i}) \textbf{Data diversity} encompasses three geometrically distinct fracture types across multiple scientific domains, including bones, ceramics, eggshells, and lithics. As shown in Fig.~\ref{fig: Dataset}, ceramics exhibit irregular, chaotic fractures typical of random breakage events, whereas flintknapping produces conchoidal fractures. This allows a systematic study of how \textit{object shapes and fracture types} affect reassembly performance. (\textbf{ii}) \textbf{Missing or extraneous fragments} are common issues in real-world scenarios~\cite{harvati2019apidima, tsesmelis2025re} (see Fig.~\ref{fig: Dataset}), which guides our model design to improve robustness.

In response to these complexities, we propose \acronym{}, a \textbf{g}eneralizable 3D re\textbf{a}ssembly framework for \textbf{r}eal-world \textbf{f}ractures, featuring four components: (\textbf{i}) \textbf{Large-scale fracture-aware pretraining} takes lessons from recent successes of large-scale pretraining in natural language and computer vision~\cite{oquab2023dinov2, radford2021learning, stevens2024bioclip, achiam2023gpt, steiner2024paligemma}. This module learns fracture segmentation of individual fragments to handle unseen objects as well as missing or extraneous fragments. (\textbf{ii}) \textbf{Flow-based reassembly on $\mathrm{SE}(3)$} introduces flow matching to learn pose distribution, leveraging the $\mathrm{SO}(3)$ manifold for accurate rotation estimation. Inspired by human puzzle-solving, we design a multi-anchor training strategy that randomly selects a subset of fragments to form local structures, exposing the model to diverse combinations to enhance distribution learning.  (\textbf{iii}) \textbf{Two-session
  flow matching at inference time} emphasizes the importance of pose initialization. The first session performs a
  one-step flow matching to generate a strong pose initialization on SE(3); the second
  session performs standard multi-step flow for refinement. (\textbf{iv}) \textbf{LoRA-based fine-tuning} enhances the model’s adaptability to specific domains.

Our main contributions are as follows:
\begin{itemize}
    \item We introduce \acronym{}, the first flow-based 3D fracture assembly framework that integrates fracture-aware pretraining and two-session
  flow matching, achieving SOTA performance across synthetic and real-world datasets. 
    \item \acronym{} sheds light on training on synthetic data for real-world challenges, effectively handling unseen object shapes and diverse fracture types, while remaining robust to missing or extraneous fragments.
    \item Collaborating with domain experts, we curate \textsc{Fractura}, a diverse dataset capturing real-world fracture complexities, to conduct the first study on 3D reassembly generalization across ceramics, bones, eggshells, and lithics. Moreover, we apply the first integration of LoRA-based fine-tuning for domain-adaptive 3D reassembly.  
\end{itemize} 

\section{\dataset{} Dataset}
\label{sec:dataset}
Existing datasets, whether synthetic~\cite{sellan2022breaking} or real~\cite{lamb2023fantastic, tsesmelis2025re}, are limited to a single fracture type. To fill this gap, we curate \dataset{}, a challenging dataset (see Fig.~\ref{fig: Dataset}), designed for a comprehensive evaluation of how \textit{object shapes}, \textit{fracture types}, and the presence of \textit{missing} or \textit{extraneous} parts affect reassembly performance.

\noindent\textbf{Data Characteristics.}
\dataset{} comprises both real and synthetic fracture subsets. In collaboration with archaeologists, paleoanthropologists, and ornithologists, the \textbf{real fracture} subset includes three real fracture types relevant to scientific challenges (see Fig.~\ref{fig: Dataset}): (i) \textit{Random breakage} produces irregular, chaotic fractures, commonly observed in \textit{ceramics}, \textit{bones}, and \textit{eggshells}. (ii) \textit{Incomplete ossification} results in unfused bone ends (epiphyses) in \textit{juvenile skeletons}, leading to fragmented rather than intact bones in skeletal collections and fossil records. Reassembling these unfused parts will benefit the following analysis and further create a complete series of bone developments over time from early childhood to adulthood. (iii) \textit{Flintnapping} produces conchoidal fractures in \textit{lithics}, characterized by radially propagating fracture lines (see Fig.~\ref{fig: Dataset}). Real-world collections and scans naturally incorporate \textit{missing or extraneous fragments}. For the \textbf{synthetic fracture} subset, we generate realistic fractures using physics-based simulation~\cite{sellan2023breaking} for \textit{ceramics}, \textit{bones}, and \textit{eggshells}, and geometry-based simulation~\cite{orellana2021proof} for \textit{lithics}.

\noindent\textbf{Data Collection and Simulation.} We utilize the high-accuracy Artec spider 3D scanner (precision: 0.05 mm) to acquire detailed 3D meshes of real fragments and intact objects from the same categories. The real fracture subset serves as test data, while intact objects are used to generate the synthetic fracture subset for fine-tuning and evaluation. Details are provided in the supplementary materials. 

\begin{table}[t]  
    \caption{\textbf{Comparisons of 3D Reassembly Datasets.} For Fracture Type*: Syn. denotes synthetic data. The number in parentheses indicates the number of synthetic/real fracture modes.}
    \label{table:Fractura} 
    \vspace{-5pt}
    \centering
    \scalebox{0.65}{
    \begin{tabular}{ccccc}
    \toprule
    \multirow{2}{*}{Datasets} & Breaking  & Fantastic  & \multirow{2}{*}{RePAIR~\cite{tsesmelis2025re}} & \multirow{2}{*}{\textbf{\dataset{}}}  \\
    & Bad~\cite{sellan2022breaking} & Breaks~\cite{lamb2023fantastic} & & \\
    
    \midrule

    \# Pieces & 8M & 300 & 1070 & 53350+292 \\
    \# Breaks & 2-100 & 2 & 2-44 & 2-22 \\
    \# Assemblies & 10474 & 150 & 117 & 9727+41 \\
    Fracture Type* & Syn. (1) & Real (1) & Real (1) & \textbf{Syn. (2) + Real (3)} \\
    \multirow{2}{*}{Object Type} & Everyday & \multirow{2}{*}{Everyday} & \multirow{2}{*}{Frescoes} & Bones, Eggshells \\
    &Artifact, Other & & & Lithics, Ceramics \\
    
    Miss. / Extra.  & $\times$ / $\times$ & $\times$ / $\times$ & $\checkmark$ / $\times$ & $\checkmark$ / $\checkmark$ \\
    Texture & $\times$ & $\times$ & $\checkmark$ & $\checkmark$ \\

    \bottomrule 
    \end{tabular}
    }
    \vspace{-10pt}
\end{table}

\noindent\textbf{Data Statistics.} Table~\ref{table:Fractura} summarizes key statistics of \dataset{} and other 3D reassembly datasets. Compared with existing datasets, \dataset{} introduces a diverse set of real and synthetic fractures across multiple scientific domains. We are actively expanding its size and diversity. More details are provided in the supplementary materials. 

\section{Method}
\label{sec:method}

Previous work either enhances global geometry learning through \textit{fragment features}~\cite{wu2023leveraging, scarpellini2024diffassemble, wang2024puzzlefusion++} or \textit{jointly} learns hierarchical features from both global and local geometry~\cite{lu2024jigsaw}. In contrast, as shown in Fig.~\ref{fig:garf}, \acronym{} decouples the understanding of \textit{local fracture} features (Sec.~\ref{Fracture-aware Pretraining}) and \textit{global} fragment alignment (Sec.~\ref{Flow-based Assembly}). At inference, we propose the two-session flow matching (Sec.~\ref{One-step Pre-assembly}) for robustness to unseen objects and increasing numbers of fractures. To further boost the performance on domain-specific data, we employ a LoRA-based fine-tuning method (Sec.~\ref{LoRA-based Fine-tuning}). 

\subsection{Why Large-Scale Fracture-Aware Pretraining?}
\label{Fracture-aware Pretraining}

\begin{figure*}[ht]
    \centering
    \includegraphics[width=\linewidth]{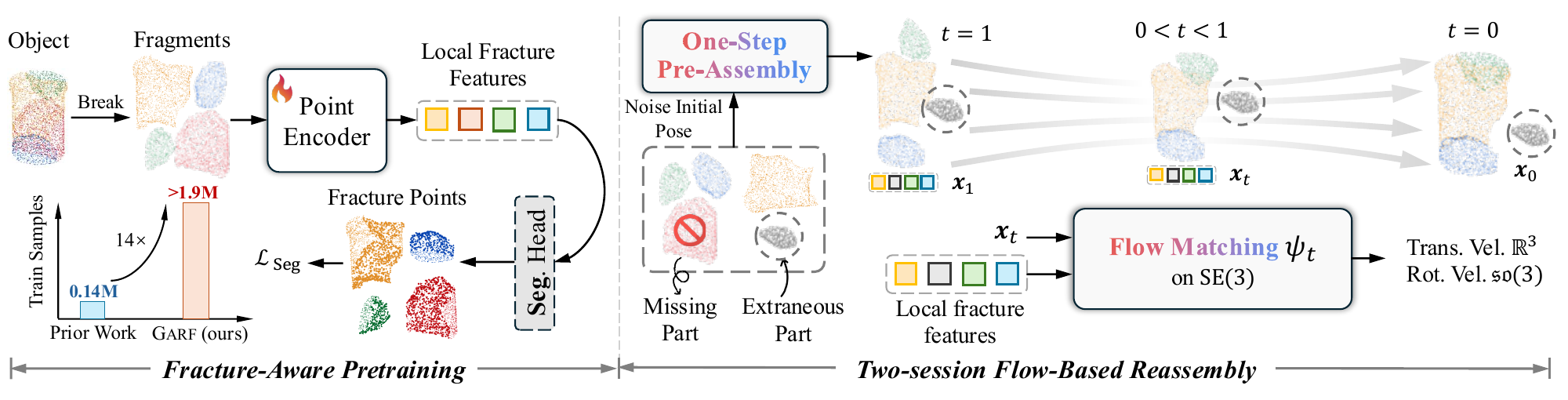}
    \vspace{-20pt}
    \caption{\textbf{Pipeline of \acronym{}.} Our framework comprises two main components: (i) Fracture-aware pretraining leverages $14\times$ more data than previous methods to learn the local fracture features via fracture point segmentation, and (ii) Two-session flow-based reassembly on $\mathrm{SE}(3)$ leverages the $\mathrm{SO}(3)$ manifold for precise rotation estimation. At inference time, one-step pre-assembly strategy provides better initial poses, enhancing robustness against unseen objects and increasing numbers of fractures.}
    \vspace{-10pt}
    \label{fig:garf}
\end{figure*}

Humans can infer fracture points without knowledge of the global object shape~\cite{papaioannou2001virtual, lu2024jigsaw}. To emulate this ability, we leverage large-scale data to learn local fracture features from individual fragments, drawing inspiration from recent advances in large-scale pretraining~\cite{oquab2023dinov2, radford2021learning, stevens2024bioclip, achiam2023gpt, steiner2024paligemma}. This enables our method to generalize to unseen object shapes and handle missing or extraneous fragments. 

Specifically, we sample a set of point clouds $\bs{P} = \{P^1, P^2, \dots, P^N\}$ to represent fragments, where $N$ is the number of fragments. To extract fracture-level features from $\bs{P}$, we employ Point Transformer V3 (PTv3)~\cite{wu2024point} as the backbone, integrating two MLP layers~\cite{lu2024jigsaw} as the segmentation head for fracture cloud segmentation. Since fracture points constitute only a small proportion of $\bs{P}$, an imbalance arises between positive and negative samples. To mitigate this, we adopt the Dice loss function~\cite{sudre2017generalised}:
\begin{equation}
\mathcal{L}_{\text{Seg}} = 1 - \frac{2\sum_{i=1}^{N}p_ig_i+\epsilon}{\sum_{i=1}^{N}p_i+\sum_{i=1}^{N}g_i+\epsilon},
\end{equation}
where $p$ is the predicted value and $g$ is its ground truth label. To derive the high-quality $g$, we directly extract shared surfaces between connected fragments from the mesh, defining them as fracture surfaces $\bs{F}$. We then apply Poisson disk sampling~\cite{bridson2007fast} to generate $\bs{P}$, ensuring that $g = \bs{P} \cap \bs{F}$. This weighted sampling based on the surface area of fragments prevents the encoder from overfitting to specific point densities, improving generalization.

\subsection{Flow-Based Reassembly on SE(3)}
\label{Flow-based Assembly}

Inspired by flow-based generative models~\cite{lipman2022flow, liu2022flow, pooladian2023multisample, lipman2024flow} in image synthesis~\cite{esser2024scaling}, protein structure generation~\cite{bose2023se, huguet2024sequence, geffnerproteina}, and robotic manipulation~\cite{black2024pi0, yim2023fast}, we leverage flow matching (FM) to model the fracture reassembly process. 

On a manifold $\mathcal{M}$, the flow $\psi_t: \mathcal{M} \rightarrow \mathcal{M}$ is defined as the solution of an ordinary differential equation (ODE):
\begin{equation}
    \frac{\mathrm{d}}{\mathrm{d}t}\psi_t(\bs{x}) = \bs{v}_t(\psi_t(\bs{x})), \quad \psi_0(\bs{x}) = \bs{x},
\end{equation}
where $\bs{v}_t(\bs{x}) \in \mathcal{T}_{\bs{x}} \mathcal{M}$ is the time-dependent vector field, and $\mathcal{T}_{\bs{x}}\mathcal{M}$ denotes the tangent space at $\bs{x}$. In the context of $\mathrm{SE}(3)$, the tangent space corresponds to the \textit{Lie algebra} $\mathfrak{se}(3)$, a six-dimensional vector space representing the velocity of the rigid motion of fragments. 

Given an object composed of $N$ fragments, we represent their poses as $\bs{T} = \{ T^1, T^2, \cdots, T^N \}$, where each $T^i$ consists of a rotation $r \in \mathrm{SO}(3)$ and a translation $a \in \mathbb{R}^3$, expressed as $T^i: \{r, a\} \in \mathrm{SE}(3)$. The initial noise distribution is defined as: $p_0(\bs{T}_0) = \mathcal{U}(\mathrm{SO}(3)) \otimes \mathcal{N}(0, I_3)$, where the rotation noise follows a uniform distribution over $\mathrm{SO}(3)$, and the translation noise is sampled from a unit Gaussian distribution. We decouple the rotation and translation flows, allowing independent flow modeling in $\mathrm{SO}(3)$ and $\mathbb{R}^3$. Therefore, the conditional flow $\bs{T}_t = \psi_t(\bs{T}_0 \mid \bs{T}_1)$ follows the geodesic path connecting $\bs{T}_0$ and $\bs{T}_1$:
\begin{equation}
    \begin{aligned}
    \bs{r}_t &= \exp_{\bs{r}_0}(t \log_{\bs{r}_0}(\bs{r}_1)), \\
    \bs{a}_t &= (1 - t) \bs{a}_0 + t \bs{a}_1,
    \end{aligned}
\end{equation}
where $\exp_{\bs{r}}$ and $\log_{\bs{r}}$ are the exponential and logarithmic maps on $\mathrm{SO}(3)$.
The final optimization objective is:
\begin{align}
\mathcal{L}_{\text{FM}} = & \mathbb{E}_{t, p_1(\bs{T}_1), p_t(\bs{T}|\bs{T}_1)}  \Biggl[ 
 \sum_{i=1}^N \Bigl\|\bs{v}^i_r(\bs{T}_t, t) - \frac{\log_{\bs{r}_t}(\bs{r}_1)}{1-t}\Bigr\|^2_g \nonumber \\
& + \Bigl\|\bs{v}^i_a(\bs{T}_t, t) - \frac{\bs{a}_1 - \bs{a}_t}{1-t}\Bigr\|^2_g \Biggr].
\end{align}
Further details are in the supplementary materials.



\noindent\textbf{Network Architecture.} Consider an object consisting of $N$ fragments with inital poses $\bs{T}_{t} \in \mathbb{R}^{N \times 7}$ at timesteps $\bs{t}$. The corresponding latent features are extracted from the pre-trained encoder: $\bs{F}=\mathcal{E}(\bs{P}) \in \mathbb{R}^{M \times c}$, where $c$ is the number of channels and $M$ is the pre-defined number of sampled points. We integrate point cloud coordinates $\bs{P}\in \mathbb{R}^{M \times 3}$, normals $\bs{n} \in \mathbb{R}^{M \times 3}$, and scale information $\bs{s}\in \mathbb{R}^{M \times 1}$ as pose-invariant shape priors in the position embedding:
\begin{equation}
    \bs{s}_{\text{emb}} = f_{\text{shape}}\Big(\operatorname{concat}\Big(\bs{F},\, \texttt{PE}(\bs{P}),\, \texttt{PE}(\bs{n}),\, \texttt{PE}(\bs{s})\Big)\Big).
\end{equation}
The pose is treated as spatial information: $\bs{p}_{\text{emb}} = f_{\text{pose}}\big(\texttt{PE}(\bs{T})\big)$. These embeddings are combined to form the Transformer input: $\bs{d} = \texttt{PE}(\bs{s}_{\text{emb}} + \bs{p}_{\text{emb}})$. 

The feature $\bs{d}$ is then processed through $L$ Transformer layers including self-attention and global attention. Leveraging the efficient computation of FlashAttention~\cite{dao2023flashattention2}, we modify the self-attention module to allow more attention on large fragments:
\begin{equation}
\begin{gathered}
\bs{Q},\, \bs{K},\, \bs{V} = \operatorname{Linear}(\operatorname{LN}(\bs{d})), \\
\bs{A}_{\text{self}} = \operatorname{FlashAttn}\Big(\bs{Q},\, \bs{K},\, \bs{V};\, \operatorname{cu}(\ell),\, \ell_{\max}\Big),
\end{gathered}
\end{equation}
where the model computes the sequence lengths $\ell$ of the variable number of sampled points per fragment batch, along with their cumulative sums $\operatorname{cu}(\ell)$. More details are in the supplementary materials. The self-attention output $\bs{A}_{\text{self}}$ is added to the original feature $\bs{d}$, yielding the updated representation $\bs{h} = \bs{d} + \bs{A}_{\text{self}}$. The global attention layer then applies a similar attention mechanism to aggregate information across fragments with $\ell=M$. The output undergoes Layer Normalization (LN) and a feed-forward network (FFN) with a residual connection, before regressing the pose $\bs{T}_{t-1}: \{\bs{r}_{t-1}, \bs{a}_{t-1}\}$ at the next timestep:
\begin{equation}
\begin{gathered}
\bs{h} \leftarrow \bs{h} + \operatorname{FFN}(\operatorname{LN}(\bs{h})), \\
\bs{a}_{t-1} = f_{\text{trans}}(\bs{h}), \quad \bs{r}_{t-1} = f_{\text{rot}}(\bs{h}). 
\end{gathered}
\end{equation}

\noindent\textbf{Multi-Anchor Training Strategy.} We observe that probability paths vary significantly across fragments; some require complex transformations, while others remain nearly stationary. To model this variation, we introduce a multi-anchor training strategy, randomly selecting $k \in [1, N-1]$ fragments as anchors and fixing their positions with identity rotations and zero translations. For these anchor fragments $\bs{i}$, the vector field is explicitly supervised to be zero: $\bs{v}^{\bs{i}}(\bs{T}_t, t) = 0$. Unlike prior approaches~\cite{wang2024puzzlefusion++}, which prevent gradient propagation for a single anchor (the largest fragment and its connected neighbors within a 50\% threshold), our multi-anchor strategy enforces a zero vector field, expanding the range of probability paths and enhancing generalization across diverse fragment configurations.


\subsection{Two-Session Flow Matching at Inference Time}
\label{One-step Pre-assembly}
Prior diffusion/flow matching work only uses single-session inference. While image generation methods use search strategies or self-supervised verifiers to improve initialization~\cite{ma2025inference, zhou2024golden, li2024enhancing, qi2024not}, such strategies are less effective for pose estimation. In contrast, we design the two-session flow matching at inference time: the first session performs a one-step FM inference ($\text{step}=1$) to generate an initial pose $\bs{T}_0'$, leveraging FM's ability to model straight-line probability paths and narrow the search space; the second
session performs standard multi-step flow for refinement. Despite a minimal 5\% increase in computational cost ($20 \rightarrow 1+20$ steps), this design outperforms single-session inference, particularly for larger fragment sets.

\begin{figure*}
    \centering
    \includegraphics[width=1.0\linewidth]{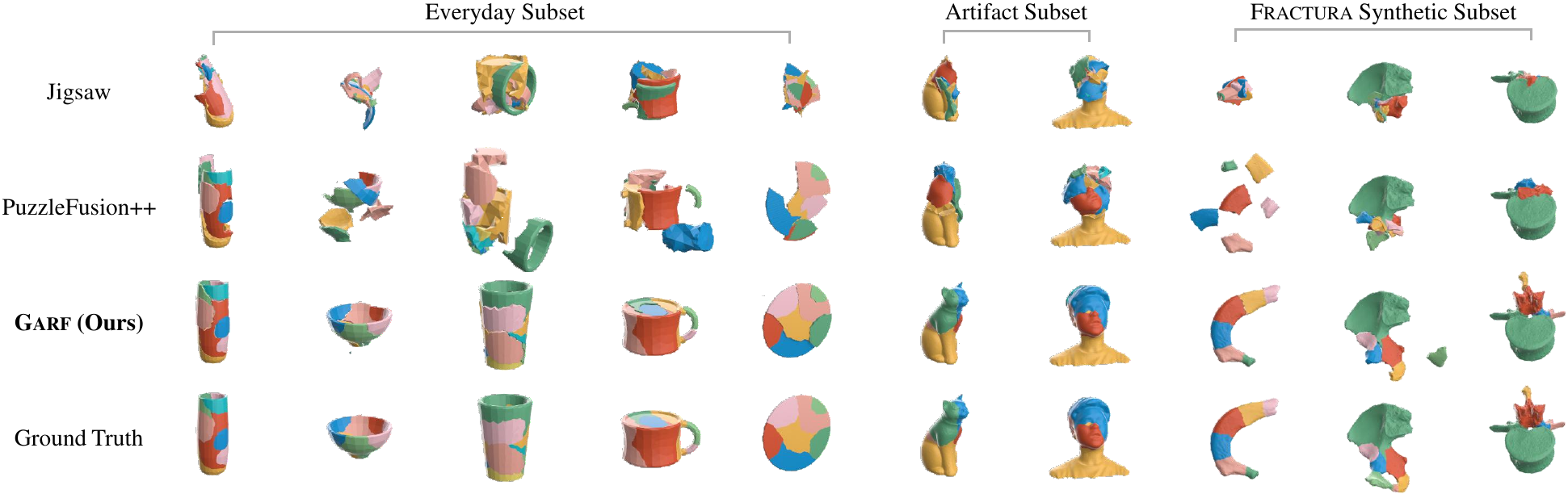}
    \vspace{-15pt}
    \caption{\textbf{Qualitative Comparisons on the Breaking Bad and \dataset{}.} \acronym{} consistently produces more accurate reassemblies, particularly on the Breaking Bad Artifact subset and \dataset{} synthetic fracture subset, demonstrating strong generalization to unseen object shapes. Meshes are used for visualization only. Additional results are available in the supplementary material.} 
    \label{fig:bb}
    \vspace{-10pt}
\end{figure*}

\subsection{LoRA-based Fine-tuning}
\label{LoRA-based Fine-tuning}
To quickly adapt to domain-specific contexts, we employ a LoRA-based~\cite{hu2022lora} fine-tuning approach using the synthetic fracture subset in \dataset{}. Specifically, we integrate LoRA adapters into the self-attention and global attention layers of the final Transformer block while unfreezing the MLP heads for pose prediction ($f_{\text{rot}}$ and $f_{\text{trans}}$). Our experiments demonstrate that this lightweight fine-tuning method requires as few as 5–10 domain-specific objects to achieve substantial improvements in scientific applications such as juvenile skeleton reconstruction and lithic refitting.


\section{Experiments}
\label{sec:exp}

\subsection{Training and Evaluation Details.}
\label{subsec:exp_setup}
\noindent
\textbf{Training.} For fracture-aware pretraining, Point Transformer V3 (PTv3)~\cite{wu2024point}
serves as the backbone, extracting 64-channel features from its final layer. \acronym{}
is pretrained on three Breaking Bad~\cite{sellan2022breaking} subsets, totaling 1.9M
fragments---$14 \times$ more than prior works. For fair comparisons, \acronym{}-mini
is pretrained only on the Everyday subset. For FM, both \acronym{} and \acronym{}-mini
are trained on the Everyday subset. We use a standard transformer~\cite{wang2024puzzlefusion++}
to compute the vector field, with each block consisting of 6 encoder layers, 8
attention heads per layer, and an embedding dimension of 512. The initial learning
rate is 2e-4 and decays by a factor of 2 at epochs 900 and 1200. \acronym{} is
trained with a batch size of 32 on 4 NVIDIA H100 HBM3 GPUs, requiring 2 days for
pretraining and 3 days for FM. \acronym{}-mini completes pretraining in 0.5 days.
For LoRA fine-tuning, we use the PEFT framework~\cite{mangrulkar2022peft} with rank
$r=12 8$, $\alpha=256$, and a dropout rate of 0.1.

\noindent
\textbf{Datasets.} We evaluate our model on three datasets with diverse object
shapes and fracture types: (\textbf{i}) \textbf{Breaking Bad}~\cite{sellan2022breaking},
the largest synthetic fracture dataset for 3D reassembly. We use the volume-constrained
version, evaluating 7,872 assemblies from the Everyday subset and 3,697
assemblies from the Artifact subset. Results on the vanilla version are in the supplementary
materials. (\textbf{ii}) \textbf{Fantastic Breaks}~\cite{lamb2023fantastic}, a real-world
dataset of 195 manually scanned fractured objects with complex surfaces, used
for evaluation only. (\textbf{iii}) \dataset{}, a mixed synthetic-real dataset
spanning ceramics, bones (vertebrae, limbs, ribs), eggshells, and lithics. For
the synthetic fracture subset, we follow an 80/20 split~\cite{tsesmelis2025re} for
LoRA fine-tuning and evaluation.

\noindent
\textbf{Evaluation Metrics.} Following~\cite{wang2024puzzlefusion++}, we evaluate
assembly quality using four metrics: (i) RMSE(R) is the root mean square error of
rotation (degrees); (ii) RMSE(T) is the root mean square error of translation; (iii)
PA is the percentage of correctly assembled fragments, where the per-fragment chamfer
distance is below 0.01; and (iv) CD is the chamfer distance between the
assembled object and ground truth.

\noindent
\textbf{Competing Methods.} We compare our approach against Global~\cite{li2020learning},
LSTM~\cite{wu2020pq}, DGL~\cite{zhan2020generative}, Jigsaw~\cite{lu2024jigsaw},
PMTR\footnote{The performance of PMTR~\cite{lee20243d} is from PuzzleFusion++~\cite{wang2024puzzlefusion++}.\label{PMTR-footlabel}}~\cite{lee20243d},
and PuzzleFusion++ (PF++) ~\cite{wang2024puzzlefusion++}. We implemented PF++~\cite{wang2024puzzlefusion++}
and Jigsaw~\cite{lu2024jigsaw} using its official codebase, while performance
metrics for other methods on the volume-constrained Breaking Bad dataset are
sourced from their official papers or repositories. Additional comparisons with SE(3)-Equiv~\cite{wu2023leveraging},
DiffAssemble~\cite{scarpellini2024diffassemble}, and PHFormer~\cite{cui2024phformer}
on the vanilla Breaking Bad dataset, as well as FragmentDiff~\cite{xu2024fragmentdiff}
on its custom Breaking Bad subset, are provided in the supplementary materials.

\subsection{Can \textbf{\acronym{}} Generalize to Unseen Shapes?}
\label{subsec:object_shape} Real-world fragmentation varies in object geometries,
requiring \acronym{} to \textit{generalize to unseen object shapes}. To
empirically assess this, we evaluate \acronym{}-mini on \dataset{} and the Artifact
subset of Breaking Bad, as well as \acronym{} on \dataset{}.

\noindent
\textbf{Results and Analysis.} Table \ref{table:BreakingBad} presents the evaluation
results on the Breaking Bad dataset, where \acronym{} achieves significant improvements
over previous SOTA methods. Compared to PF++~\cite{wang2024puzzlefusion++}, \acronym{}
reduces rotation error by 82.87\% and translation error by 79.83\%, while achieving
a CD below 0.001, indicating near-perfect reconstructions. Even more impressively,
\acronym{}-mini demonstrates exceptional generalization capability. Despite
being trained only on the Everyday subset, it maintains consistent performance on
the \textit{unseen} Artifact subset, avoiding the performance degradation typically
observed with unseen object shapes. This highlights \acronym{}'s robust feature extraction
and assembly mechanisms. Notably, \acronym{} achieves 95.33\% and 95.04\% PA on
the Breaking Bad dataset, approaching the theoretical maximum PA of 96.49\% (Everyday
subset) and 96.10\% (Artifact subset) when excluding fragments smaller than 0.1\%
of the object volume\footnote{Scanning tiny 3D fragments in real-world settings
is inherently challenging; anthropologists often treat such fragments as missing
parts.\label{garf-footlabel}}. On the challenging synthetic subset of \dataset{},
which contains \textit{unseen domain-specific objects}, \acronym{} further demonstrates
superior generalization capability, outperforming all competing methods across
all evaluation metrics. As shown in Fig.~\ref{fig:bb}, \acronym{} consistently
produces more accurate reassemblies, particularly on the Breaking Bad Artifact subset
and the \dataset{} synthetic fracture subset, confirming its strong generalization
to unseen object shapes.


\begin{table}[t]
    \caption{\textbf{Quantitative Results on Volume-Constrained Breaking Bad~\cite{sellan2022breaking}
    and \dataset{} Datasets.} The best performance metric is highlighted in \textbf{bold},
    while the second-best is \underline{underlined}.}
    \label{table:BreakingBad}
    \vspace{-5pt}
    \centering
    \scalebox{0.85}{
    \begin{tabular}{ccccc}
        \toprule \multirow{2}{*}{Methods}                                                                       & RMSE(R) $\downarrow$ & RMSE(T) $\downarrow$ & PA $\uparrow$     & CD $\downarrow$  \\
                                                                                                                & \text{degree}        & $\times 10^{-2}$     & \%                & $\times 10^{-3}$ \\
        \midrule \multicolumn{5}{>{\columncolor{green!10}}c}{Tested on the \textbf{Everyday} Subset}             \\
        \midrule Global~\cite{li2020learning}                                                                   & 80.50                & 14.60                & 28.70             & 13.00            \\
        LSTM~\cite{wu2020pq}                                                                                    & 82.70                & 15.10                & 27.50             & 13.30            \\
        DGL~\cite{zhan2020generative}                                                                           & 80.30                & 13.90                & 31.60             & 11.80            \\
        Jigsaw~\cite{lu2024jigsaw}                                                                              & 42.19                & 6.85                 & 68.89             & 8.22             \\
        PMTR~\cite{lee20243d}                                                                                   & 31.57                & 9.95                 & 70.60             & 5.56             \\
        PF++~\cite{wang2024puzzlefusion++}                                                                      & 35.61                & 6.05                 & 76.17             & 2.78             \\
        \textbf{\acronym{}-mini }                                                                               & \underline{6.68}     & \underline{1.34}     & \underline{94.77} & \underline{0.25} \\
        \textbf{\acronym{} }                                                                                    & \textbf{6.10}        & \textbf{1.22}        & \textbf{95.33}    & \textbf{0.22}    \\
        \midrule \multicolumn{5}{>{\columncolor{green!10}}c}{Tested on the \textbf{Artifact} Subset}             \\
        \midrule Jigsaw                                                                                         & 43.75                & 7.91                 & 65.12             & 8.50             \\
        PF++                                                                                                    & 47.03                & 10.63                & 57.97             & 8.24             \\
        \textbf{\acronym{}-mini }                                                                               & \underline{7.67}     & \underline{1.77}     & \underline{93.34} & \underline{0.81} \\
        \textbf{\acronym{} }                                                                                    & \textbf{5.82}        & \textbf{1.27}        & \textbf{95.04}    & \textbf{0.42}    \\
        \midrule \multicolumn{5}{>{\columncolor{red!10}}c}{Tested on the \textbf{Fractura} (Synthetic Fracture)} \\
        \midrule Jigsaw                                                                                         & 60.50                & 18.49                & 33.06             & 70.68            \\
        PF++                                                                                                    & 62.57                & 18.65                & 37.74             & 36.13            \\
        \textbf{\acronym{}-mini }                                                                               & \underline{27.88}    & \underline{6.79}     & \underline{76.25} & \underline{7.70} \\
        \textbf{\acronym{} }                                                                                    & \textbf{19.63}       & \textbf{4.93}        & \textbf{83.41}    & \textbf{6.06}    \\
        \bottomrule
    \end{tabular}}
    \vspace{-5pt}
\end{table}

\subsection{How Do Fracture Types Affect Generalization?}
\label{subsec:fracture_type} As scanning time significantly increases with fragment
count~\cite{lamb2023fantastic}, real fracture datasets are typically limited in
\textit{size} and \textit{diversity}, making large-scale training infeasible. To
address this, we investigate \textit{how fracture types impact zero-shot
generalization} on Fantastic Breaks and \dataset{}, as well as \textit{the fine-tuning
performance using domain-specific synthetic fractures} on \dataset{}.

\noindent
\textbf{Real Fracture on Fantastic Breaks~\cite{lamb2023fantastic}.} As shown in
Table~\ref{table:FantasticBreaks}, \acronym{} demonstrates superior generalization
to real-world fracture surfaces, achieving a remarkable 48.65\% reduction in rotation
error compared to PF++~\cite{wang2024puzzlefusion++}, indicating that our model
effectively bridges the synthetic-to-real gap in fracture surface understanding
for everyday objects.

\noindent
\textbf{Real Fracture on \dataset{}.} We further evaluate performance across three
fracture types. To isolate the impact of unseen objects, $\acronym{}_{\text{LoRA}}$
is fine-tuned separately on synthetic fractures from bones, eggshells, and lithics
in \dataset{}\footnote{Fine-tuning is not performed for ceramics, as its object categories
closely resemble those in the Everyday subset of Breaking Bad.\label{ceramics-footlabel}}.
Figure~\ref{fig:real_fracture} compares reassembly performance across three
fracture types in the \dataset{} real fracture subset. \acronym{} outperforms competing
methods on ceramics even with three missing fragments.
$\acronym{}_{\text{LoRA}}$ further improves performance by mitigating the effect
of unseen objects. $\acronym{}_{\text{LoRA}}$ generalize well to random breakage
(limb bones, ceramics) and incomplete ossification (vertebrae). Unfortunately,
although finetune significantly improves the performance on lithics, both \acronym{}
and $\acronym{}_{\text{LoRA}}$ struggle on lithics, likely due to high ambiguity
among flakes and the core, a well-known challenging spatial reasoning problem
for anthropologists~\cite{laughlin_experimental_2010}. This unresolved challenge
in \dataset{} offer valuable directions for future research.

\begin{table}[t]
    \caption{\textbf{Quantitative Results on Fantastic Breaks Dataset~\cite{lamb2023fantastic}}.
    This includes manually collected real-world objects.}
    \label{table:FantasticBreaks}
    \vspace{-5pt}
    \centering
    \scalebox{0.85}{
    \begin{tabular}{ccccc}
        \toprule \multirow{2}{*}{Methods}                                                        & RMSE(R) $\downarrow$ & RMSE(T) $\downarrow$ & PA $\uparrow$     & CD $\downarrow$  \\
                                                                                                 & degree               & $\times 10^{-2}$     & \%                & $\times 10^{-3}$ \\
        \midrule \rowcolor{yellow!20}\multicolumn{5}{c}{Tested on the \textbf{Fantastic Breaks} } \\
        \midrule Jigsaw~\cite{lu2024jigsaw}                                                      & 26.30                & 6.43                 & 73.64             & 10.47            \\
        PF++~\cite{wang2024puzzlefusion++}                                                       & \underline{20.68}    & \underline{4.37}     & \underline{83.33} & \underline{6.68} \\
        \textbf{\acronym{}}                                                                      & \textbf{10.62}       & \textbf{2.10}        & \textbf{91.00}    & \textbf{2.12}    \\
        \bottomrule
    \end{tabular}
    }
    \vspace{-10pt}
\end{table}
\begin{figure}
    \centering
    \includegraphics[width=1\linewidth]{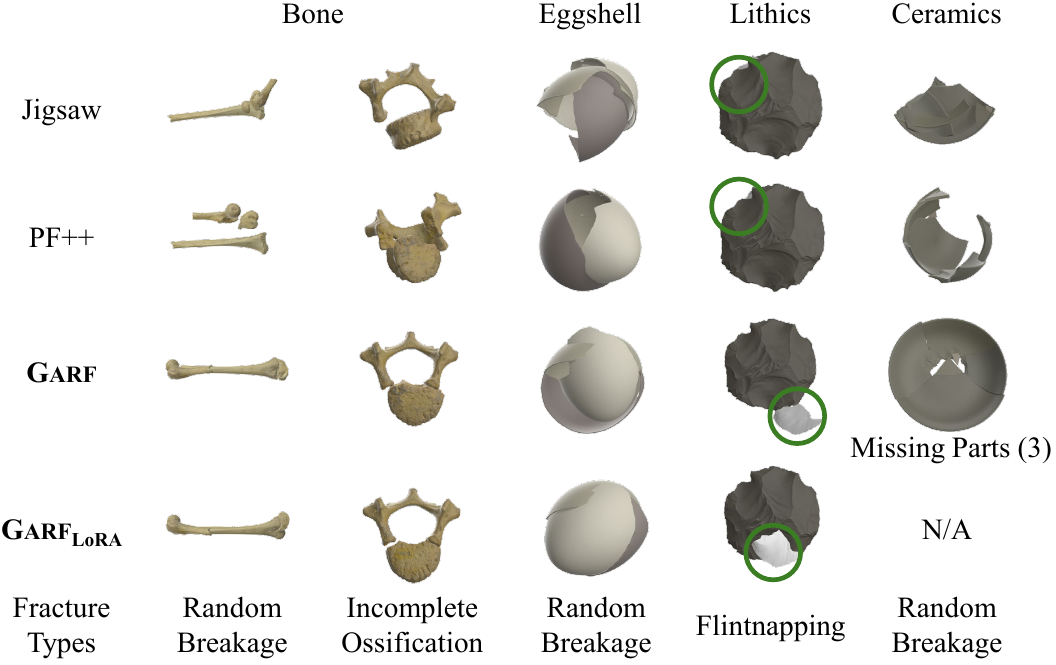}
    \caption{\textbf{Qualitative Comparisons on the \dataset{} real fracture subset.}
    \acronym{} generalizes well to random breakage (limb bones and ceramics) and
    incomplete ossification (vertebrae) but faces challenges with high-ambiguity
    fractures like flintknapping (lithics). Fine-tuning enhances performance,
    particularly for thin-shell structures (eggshells) and flintnapping (lithics).}
    \label{fig:real_fracture}
    \vspace{-16pt}
\end{figure}

\subsection{How Do Missing or Extraneous Parts Affect Performance?}
\label{subsec:missing_part}

Archaeological materials are often \textit{incomplete} or mixed with similar but
\textit{extraneous} fragments, posing significant challenges for assembly. To
quantitatively assess model robustness under these conditions at scale, we extend
Breaking Bad's Everyday subset in two ways: (i) \textit{Missing} parts subset removes
20\% of fragments in descending order of volume, preserving the largest anchor fragments
and maintaining the object's connectivity graph; (ii) \textit{Extraneous} parts subset
adds fragments from other objects in the same category, selecting pieces smaller
than the largest anchor fragment but larger than 5\% of the object's total volume
to ensure they are not trivially small. For a fair comparison, we evaluate objects
with 5 to 17 fragments. Additionally, we provide visual demonstrations of assemblies
with naturally missing or extraneous parts in \dataset{}.

\noindent
\textbf{Results and Analysis.} Table~\ref{table:missing} shows that \acronym{}
demonstrates strong resilience, with minimal performance degradation over competing
methods. With 20\% extraneous fragments, \acronym{} maintains a high PA of 79.21\%,
only 5.0\% lower than with complete sets, whereas PF++ drops by 24.32\%. Similarly,
with 20\% missing fragments, \acronym{} achieves 78.87\% PA, far surpassing Jigsaw
(28.85\%) and PF++ (47.22\%). Figure~\ref{fig:missing} illustrates how missing
or extraneous fragments affect reassembly performance. While all methods degrade
under these challenging conditions, \acronym{} demonstrates superior robustness,
maintaining coherent structures despite missing or extraneous fragments. In contrast,
Jigsaw and PF++ exhibit severe misalignments and fragment mismatches. This robustness
suggests that \acronym{} can partially handle missing or extraneous fragments, benefiting
from our model design. 

\begin{table}[t]
    \caption{\textbf{Quantitative Results on Missing / Extraneous Parts.} }
    \label{table:missing}
    \vspace{-5pt}
    \centering
    \scalebox{0.75}{
    \begin{tabular}{cccccc}
        \toprule Methods                                              & Input                & RMSE(R) $\downarrow$ & RMSE(T) $\downarrow$ & PA $\uparrow$  & CD $\downarrow$ \\
        \midrule

\multirow{3}{*}{Jigsaw~\cite{lu2024jigsaw}}         & Complete             & 71.41                & 15.83                & 28.34          & 21.94           \\
                                                                      & 20\% Miss.           & 70.45                & 15.54                & 28.85          & 21.58           \\
                                                                      & 20\% Extra.          & 74.55                & 19.02                & 24.03          & 26.15           \\
        \midrule

\multirow{3}{*}{PF++~\cite{wang2024puzzlefusion++}} & Complete             & 59.26                & 12.00                & 49.38          & 5.52            \\
                                                                      & 20\% Miss.           & 61.14                & 12.26                & 44.71          & 7.57            \\
                                                                      & 20\% Extra.          & 61.25                & 13.91                & 40.59          & 8.77            \\
        \midrule

\multirow{3}{*}{\textbf{\acronym{}}}                & Complete             & 19.55                & 3.83                 & 83.39          & 0.62            \\
                                                                      & \textbf{20\% Miss.}  & \textbf{22.23}       & \textbf{4.75}        & \textbf{78.87} & \textbf{1.40}   \\
                                                                      & \textbf{20\% Extra.} & \textbf{22.70}       & \textbf{4.62}        & \textbf{79.21} & \textbf{1.29}   \\
        \bottomrule
    \end{tabular}
    }
    \vspace{-5pt}
\end{table}

\begin{figure}[t]
    \centering
    \includegraphics[width=1\linewidth]{
        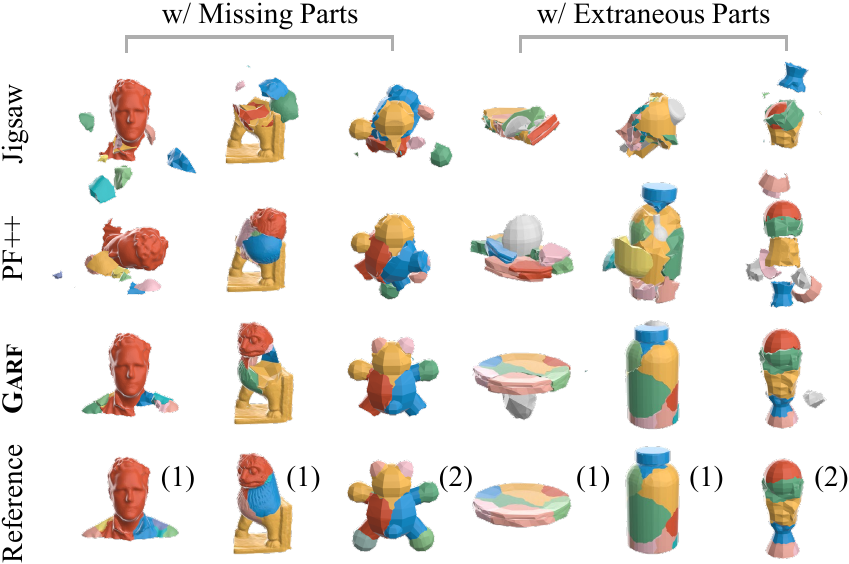
    }
    \vspace{-15pt}
    \caption{\textbf{Qualitative Comparisons on the Missing or Extraneous Impact.}
    \acronym{} demonstrates superior robustness, maintaining coherent structures
    despite missing or extraneous fragments.}
    \label{fig:missing}
\end{figure}

\subsection{Ablation Study}
\label{subsec:ablation}


\noindent
\textbf{Designs of Model.} We incrementally add our key design choices to the PF++~\cite{wang2024puzzlefusion++}
baseline, reporting RMSE (R/T) and PA as shown in Table~\ref{table:model_designs}.
\bcircled{1} \textbf{Fracture-aware Pretraining Strategy.} We replace the VAE-based
pretraining used in PF++ with our fracture-aware pretraining strategy. Our
strategy reduces RMSE(R) by 53.3\% and RMSE(T) by 49.0\%. \bcircled{2} \textbf{More
attention on large fragments.} We observe that large fragments are easier to assemble
and yield more stable gradients. To leverage this, we apply surface-area-based weighted
sampling and modify the attention mechanism to emphasize large fragments. This
leads to substantial improvements. \textbf{\bcircled{3} Two-Session Flow Matching
on SE(3).} We design a two-session flow matching scheme: the first session performs
one-step coarse pose initialization, while the second refines it via multi-step
flow. This design further boosts performance during inference. More experimental results and analysis are in the supplementary materials.


\begin{table}[h]
    \caption{\textbf{Ablation Study on Our Designs of Model.} }
    \label{table:model_designs}
    \vspace{-5pt}
    \centering
    \scalebox{0.68}{
    \begin{tabular}{lcccccc}
        \toprule Setups                                      & Backbone              & Denoiser            & RMSE(R) $\downarrow$  & RMSE(T) $\downarrow$  & PA $\uparrow$          \\
        \midrule I: PuzzleFusion++                           & PointNet++            & Diffusion           & 32.91                 & 5.26                  & 78.95                  \\
        I + \bcircled{1}                                     & PointNet++            & Diffusion           & 15.36                 & 2.68                  & 89.40                  \\
        I + \bcircled{1}                                     & PointNet++            & FM                  & 15.55                 & 2.69                  & 89.43                  \\
        I + \bcircled{1} \bcircled{2}                        & PointNet++            & Diffusion           & 12.58                 & 2.16                  & 91.24                  \\
        I + \bcircled{1} \bcircled{2}                        & PointNet++            & FM                  & 8.70                  & 1.67                  & 93.60                  \\
        I + \bcircled{1} \bcircled{2} \bcircled{3}           & PointNet++            & FM                  & 7.40                  & 1.47                  & 94.43                  \\
        \midrule

I + \bcircled{1} \bcircled{2} \bcircled{3} & \multirow{2}{*}{PTv3} & \multirow{2}{*}{FM} & \multirow{2}{*}{6.68} & \multirow{2}{*}{1.34} & \multirow{2}{*}{94.77} \\
        (GARF-mini)                                          &                       &                     &                       &                       &                        \\
        \bottomrule
    \end{tabular}
    }
\end{table}

\begin{figure}[t]
    \centering
    \includegraphics[width=\linewidth]{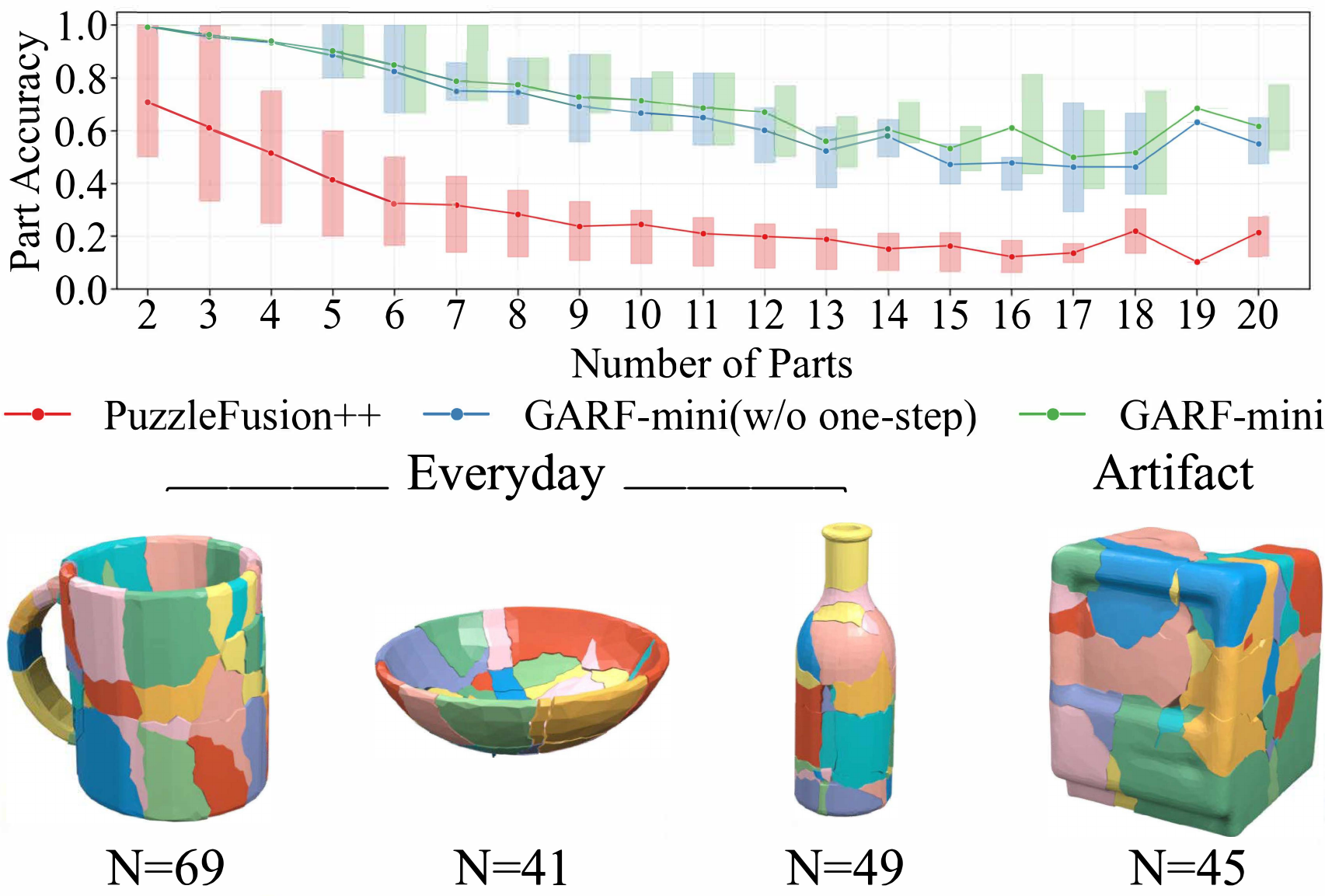}
    \caption{\textbf{Investigation on the Number of Fragments.} \acronym{}-mini
    significantly outperforms PF++, with the two-session FM further boosting results
    for $> 10$ fragments. Notably, \acronym{}-mini also generalizes to fractures
    containing well over 20 fragments.}
    \label{fig:number}
    \vspace{-15pt}
\end{figure}

\noindent
\textbf{Number of Fragments.} We analyze performance across varying fragment counts,
as shown in Figure~\ref{fig:number}. \acronym{}-mini consistently surpasses PF++
across all fragment counts. Our two-session FM further enhances performance on unseen
objects containing more than 10 fragments. Interestingly, although \acronym{}
has only been trained on data with 2-20 fragments in Breaking Bad, it can still generalize
to fractures with more than 20 pieces as shown in Figure~\ref{fig:number}.
\section{Impact and Limitations}
\label{sec:future}

\noindent\textbf{Scientific Impacts.} How objects—whether they are bones, ceramic pots, or stone tools—were reassembled and what processes influenced their reconstruction is one of the basic questions common to different research communities, including \textit{paleontology} and \textit{paleoanthropology}, \textit{archaeology}, and \textit{forensic science}. To explore this, we make the first attempt to collaborate with archaeologists, paleoanthropologists, and ornithologists to build a generalizable model for real-world fracture reassembly. While \acronym{} achieves significant improvements, challenges remain, particularly in handling the complexities presented by \dataset{}, which creates new opportunities for the vision and learning communities, encouraging advancements in 3D puzzle solving. 

\noindent\textbf{Limitations and Future Directions.} 
\acronym{} encounters challenges in handling fracture ambiguity, especially when fragments have subtle geometric differences or inherently ambiguous fracture surfaces. For instance, it struggles with lithic refitting in \dataset{} due to high ambiguity among flakes and the core, as well as fresco reconstruction in RePAIR, where erosion affects fracture surfaces~\cite{tsesmelis2025re}. Therefore, our future work will focus on: (i) Multimodal fracture reassembly, integrating geometric and texture information; (ii) Test-time policy optimization using expert feedback; (iii) Expanding the size and diversity of \dataset{}.

\section{Related Work}
\label{sec:related}

\noindent
\textbf{Fracture Assembly.} Early methods relied on explicit geometric matching
with handcrafted features~\cite{luo2013co, andreadis2014facet, xu2015robust, holland2022digital},
often struggling with ambiguous or incomplete geometries. The advent of the
large-scale synthetic dataset Breaking Bad~\cite{sellan2022breaking} has enabled
learning-based approaches to acquire robust geometric representations and
assembly strategies~\cite{lu2024survey}. Jigsaw~\cite{lu2024jigsaw} jointly learns
hierarchical features from global and local geometries for fracture matching and
pose estimation, while SE(3)-equiv~\cite{wu2023leveraging} extracts fragment features
for pose estimation. DiffAssemble~\cite{scarpellini2024diffassemble} improves
performance using a diffusion model. PuzzleFusion++~\cite{wang2024puzzlefusion++}
mimics how humans solve spatial puzzles by integrating diffusion-based pose estimation
with a VAE-based fragment representation and transformer-based alignment verification.
However, while PuzzleFusion++ achieves SOTA results on the everyday subset, its
performance degrades significantly on the Artifact subset~\cite{wang2024puzzlefusion++}.
More critically, it remains unclear whether models trained on synthetic data can
generalize to real-world fractures with more complex breakage patterns. To fill this
gap, we identify major real-world fracture challenges and curate \dataset{}, a
dataset capturing key complexities. To address these challenges, we propose
\acronym{}, a generalizable 3D reassembly framework for real-world fractures.

\noindent
\textbf{Flow Matching.}
Recent advances have explored flow matching (FM) across image generation~\cite{esser2024scaling},
protein backbone generation~\cite{yim2023fast, huguet2024sequence,
geffnerproteina}, and general robot control~\cite{black2024pi0}. AssembleFlow~\cite{guo2025assembleflow}
attempts to leverage FM for molecular assembly, but introduces numerical errors
by approximating quaternion updates through direct addition over small time
intervals during inference. While diffusion models have been widely applied to fracture
assembly~\cite{scarpellini2024diffassemble, wang2024puzzlefusion++}, FM provides
a more natural formulation by learning geodesic flows in SE(3). We propose the
first FM-based fracture assembly framework, incorporating the multi-anchor training
strategy and two-session flow matching at inference time.

\section{Conclusion}
\label{sec:conclusion}
In collaboration with archaeologists, paleoanthropologists, and ornithologists, we present \dataset{}, a diverse and challenging fracture assembly dataset, and \acronym{}, a generalizable 3D reassembly framework designed for real-world fractures. \dataset{} serves as a challenging benchmark to evaluate how object shapes, fracture types, and the presence of missing or extraneous parts affect reassembly performance.
Facing these challenges, \acronym{} offers vital guidance on training on synthetic data to advance real-world 3D puzzle solving. Comprehensive evaluations demonstrate its strong generalization to unseen object shapes and diverse fracture types. Despite its superior performance, \acronym{} still struggles with geometric ambiguity, particularly when dealing with highly similar fragments and eroded fracture surfaces. We anticipate that \dataset{} will drive further advancements in 3D reassembly, pushing the boundaries of spatial reasoning to answer unknown scientific questions.


\noindent\textbf{Acknowledgement.} We gratefully acknowledge the Physical Anthropology Unit, Universidad Complutense de Madrid for access to curated human skeletons, and Dr. Scott A. Williams (NYU Anthropology Department) for the processed data samples. This work was supported in part through NSF grants 2152565, 2238968, 2322242, and 2426993, and the NYU IT High Performance Computing resources, services, and staff expertise.

{
    \small
    \bibliographystyle{ieeenat_fullname}
    \bibliography{main}
}

\newpage
\clearpage
\setcounter{page}{1}

\section*{Appendix}
\label{sec:appendix}
\renewcommand{\thesection}{\Alph{section}}
\renewcommand{\thefigure}{\Roman{figure}}
\renewcommand{\thetable}{\Roman{table}}

\setcounter{section}{0}
\setcounter{figure}{0}
\setcounter{table}{0}

This document supplements the main paper as follows:

\renewcommand{\labelenumii}{\Roman{enumii}}

\begin{enumerate}
    \item Dataset details (Section~\ref{sec:dataset}).

    \item More details about the training recipe and reproducibility (section~\ref{sec:method}).

    \item More visualizations and detailed tables (section~\ref{sec:exp}).
\end{enumerate}

\section{Additional Dataset Details}
\label{sec:dataset}
\subsection{Fracture Simulation}
(i) \textit{Bone}. For elongated structures like limbs and ribs, we used Blender's
skinning and subdivision surface techniques to create realistic cylindrical
hollows, replicating bone morphology. We then applied the physics-based fracture
method from Breaking Bad~\cite{sellan2022breaking} to generate 2–20 fragments. The
same approach was used for \textit{os coxae} and vertebrae, forming the simulated
subset of the bone category. (ii) \textit{Eggshell}. Since scanned eggshells
produce watertight solid ellipsoids, we removed 98\% of the concentric volume to
simulate thin shells. We then applied the same physics-based fracture method to generate
realistic breakage patterns. (iii) \textit{Ceramics}. Given that ceramic objects
(e.g., bowls, pots, vases) closely resemble those in Breaking Bad's everyday
category, we focused on scanning real fragments and did not include a simulated subset.
(iv) \textit{Lithics}. As an initial feasibility test, two generalized core morphologies
were repeatedly virtually knapped with some randomized variation following methods
described for the dataset in~\cite{orellana2021proof} to produce core and flake combinations
with varying geometries.

\subsection{\textbf{\dataset{}} Statistics}
Table~\ref{table:data_statistics} presents detailed statistics for each category
in \dataset{}. We continue to expand both the dataset’s scale and diversity, aiming
to establish a comprehensive cyberinfrastructure for the vision-for-science community.
\begin{table}[ht]
    \caption{Dataset Statistics of the \dataset{} Dataset.}
    \label{table:data_statistics}
    \centering
    \scalebox{0.80}{
    \begin{tabular}{cccc}
        \toprule \textbf{Category}               & \textbf{Fracture Type} & \textbf{\# Assemblies} & \textbf{\# Pieces} \\
        \midrule \multirow{2}{*}{Bone}           & Real                   & 17                     & 37                 \\
                                                 & Synthetic              & 7056                   & 39943              \\
        \midrule \multirow{2}{*}{Eggshell}       & Real                   & 3                      & 12                 \\
                                                 & Synthetic              & 2268                   & 12600              \\
        \midrule \multirow{2}{*}{Ceramics}       & Real                   & 9                      & 51                 \\
                                                 & Synthetic              & N/A                    & N/A                \\
        \midrule \multirow{2}{*}{Lithics}        & Real                   & 12                     & 192                \\
                                                 & Synthetic              & 403                    & 807                \\
        \midrule \multirow{2}{*}{\textbf{Total}} & Real                   & 41                     & 292                \\
                                                 & Synthetic              & 9727                   & 53350              \\
        \bottomrule
    \end{tabular}
    }
\end{table}

\section{Additional Implementation Details}

\label{sec:method}
\subsection{Data Preprocessing}
We preprocess the BreakingBad dataset~\cite{sellan2022breaking} to calculate the
segmentation ground truth directly from meshes to reduce the computation
overhead during training as described in Sec.~\ref{Fracture-aware Pretraining},
and there's no need for any hyperparameters. Unlike baseline methods (Global, LSTM,
and DGL) provided by the dataset and PF++~\cite{wang2024puzzlefusion++}, which
samples $M=1000$ points from the mesh per fragment, we used the same setting as in
Jigsaw~\cite{lu2024jigsaw} to sample $M=5000$ points per object, making all
fragments have the same point density. With this sampling setting, we did not
encounter any gradient explosion issues during training, as reported in FragmentDiff~\cite{xu2024fragmentdiff},
which occur when sampling too many points for tiny pieces. Meanwhile, we employ
the Poisson disk sampling method to ensure that the points are more uniformly
distributed on the surface of the fragment. During training, standard data
augmentation techniques are applied, including recentering, scaling, and random rotation.

\subsection{Training Recipe}

We modified a smaller version of Point Transformer V3~\cite{wu2024point} as our
backbone for the segmentation pretraining, as shown in Table~\ref{table:training_recipe},
which we found to be sufficient and more memory efficient. Since \acronym{} uses
a much larger training dataset, we reduce the training epochs to 150, other than
the 400 epochs used in \acronym{}-mini. Both pretrainings reach over 99.5\% accuracy
on the validation set. Samples of segmentation results are shown in Fig.~\ref{fig:segmentation}.

\begin{figure}[h]
    \centering
    \includegraphics[width=0.7\linewidth]{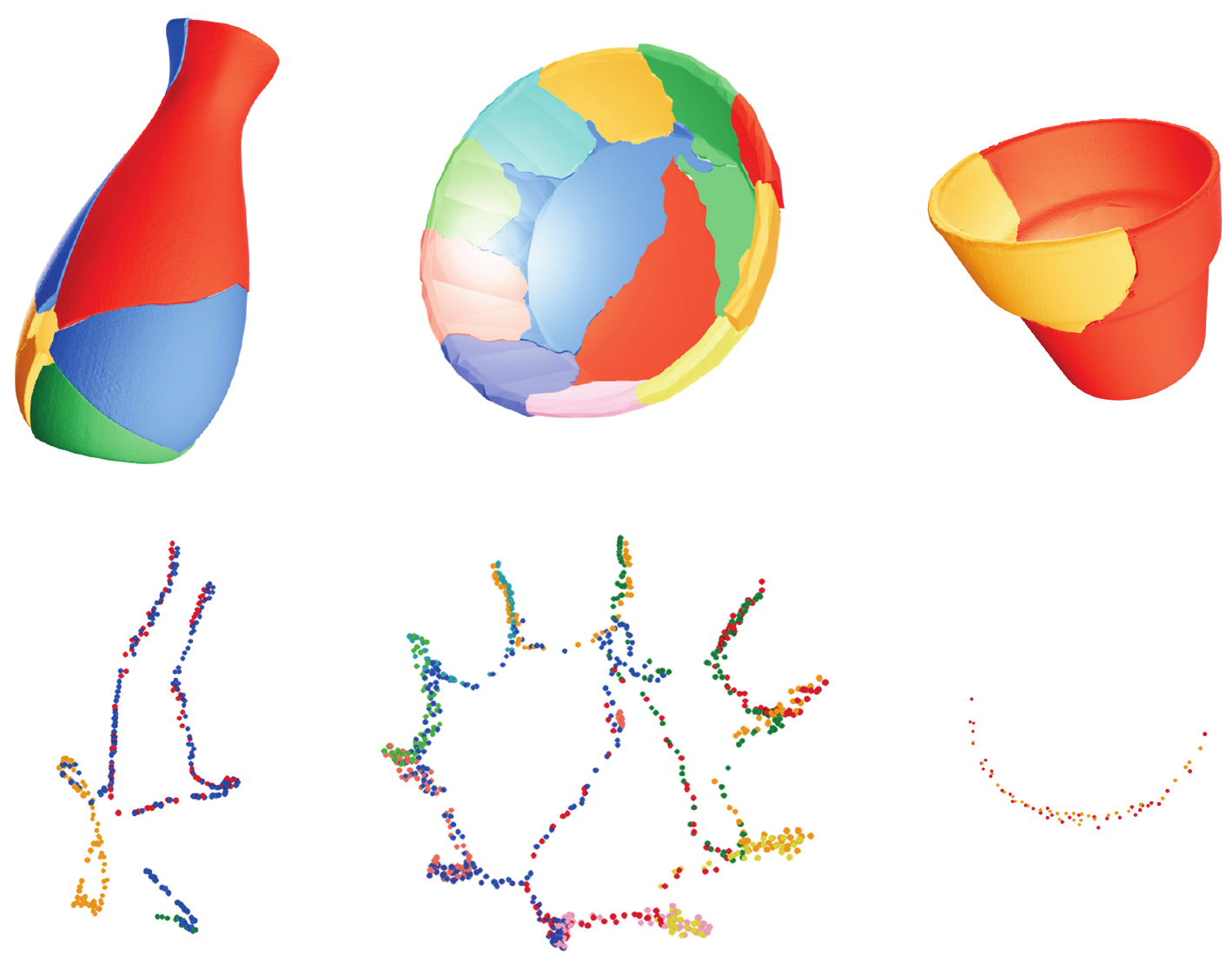}
    \caption{Segmentation results on a real-world object (left), Breaking Bad~\cite{sellan2022breaking}
    (center) and Fantastic Breaks~\cite{lamb2023fantastic} (right).}
    \label{fig:segmentation}
\end{figure}

For FM training, we provide the hyperparameters in Table~\ref{table:training_recipe}
for reproducibility. The settings are identical for both \acronym{} and \acronym{}-mini,
as their only difference lies in the pretraining stage.

\begin{table}[t]
    \caption{Training Configurations.}
    \label{table:training_recipe}
    \centering
    \scalebox{0.70}{
    \begin{tabular}{cll}
        \toprule                             & \multicolumn{1}{c}{\textbf{Config}} & \multicolumn{1}{c}{\textbf{Value}} \\
        \midrule \multirow{8}{*}{Backbone}   & Encoder Depth                       & [2, 2, 6, 2]                       \\
                                             & Encoder \# Heads                    & [2, 4, 8, 16]                      \\
                                             & Encoder Patch Size                  & [1024, 1024, 1024, 1024]           \\
                                             & Encoder Channels                    & [32, 64, 128, 256]                 \\
                                             & Decoder Depth                       & [2, 2, 2]                          \\
                                             & Decoder \# Heads                    & [4, 8, 16]                         \\
                                             & Decoder Patch Size                  & [1024, 1024, 1024]                 \\
                                             & Decoder Channels                    & [256, 128, 64]                     \\
        \midrule\multirow{6}{*}{Pretraining} & Global Batch Size                   & 256                                \\
                                             & Epochs                              & 400 / 150                          \\
                                             & Learning Rate                       & 1e-4                               \\
                                             & Scheduler                           & CosineAnnealingWarmRestarts        \\
                                             & Scheduler $T_{0}$                   & 100 / 50                           \\
                                             & \# Trainable Params                 & 12.7M                              \\
        \midrule\multirow{7}{*}{Training}    & Global Batch Size                   & 128                                \\
                                             & Epochs                              & 1500                               \\
                                             & Learning Rate                       & 2e-4                               \\
                                             & Scheduler                           & MultiStepLR                        \\
                                             & Scheduler Milestones                & [900, 1200]                        \\
                                             & Scheduler $\gamma$                  & 0.5                                \\
                                             & \# Trainable Params                 & 43.5M                              \\
        \bottomrule
    \end{tabular}
    }
\end{table}

\subsection{Preliminaries on Riemannian Flow Matching}
Instead of simulating discrete noise addition steps, flow matching (FM) learns a
probability density path $p_{t}$, which progressively transforms a noise
distribution $p_{t=0}$ to the data distribution $p_{t=1}$, with a time variable $t
\in [0, 1]$. As a simulation-free method aiming to learn continuous normalizing
flow (CNF), FM models a probability density path $p_{t}$, which progressively
transforms a noise distribution $p_{t=0}$ to the data distribution $p_{t=1}$, with
a time variable $t \in [0, 1]$. Inspired by \textit{learning assembly by
breaking}, the rigid motion of the fragments corresponds to the geodesic on the
\textit{Lie group} $\mathrm{SE}(3)$, which is a differentiable Riemannian
manifold. Inspired by previous works~\cite{bose2023se, yim2023fast,
geffnerproteina}, FM can be extended to $\mathrm{SE}(3)$ manifold to learn the
rigid assembly process.

On a manifold $\mathcal{M}$, the flow $\psi_{t}: \mathcal{M}\rightarrow \mathcal{M}$
is defined as the solution of an ordinary differential equation (ODE):
\begin{equation}
    \frac{\mathrm{d}}{\mathrm{d}t}\psi_{t}(\bs{x}) = \bs{v}_{t}(\psi_{t}(\bs{x}))
    , \quad \psi_{0}(\bs{x}) = \bs{x},
\end{equation}
where $\bs{v}_{t}(\bs{x}) \in \mathcal{T}_{\bs{x}}\mathcal{M}$ is the time-dependent
vector field, and $\mathcal{T}_{\bs{x}}\mathcal{M}$ is the tangent space of the manifold
at $\bs{x}\in \mathcal{M}$. In the context of $\mathrm{SE}(3)$, the tangent space
is the \textit{Lie algebra} $\mathfrak{se}(3)$, which is a six-dimensional vector
space, presenting the velocity of the rigid motion of the fragments. Given the \textit{conditional
vector filed}
$\bs{u}_{t}(\bs{x}\mid \bs{x}_{1}) \in \mathcal{T}_{\bs{x}}\mathcal{M}$, which generates
the conditional probability path $p_{t}(\bs{x}\mid \bs{x}_{1})$, the Riemannian
flow matching objective can be defined as:
\begin{equation}
    \mathcal{L}_{\text{CFM}}:= \mathbb{E}_{t, p_1(\bs{x}_1), p_t(\bs{x} \mid \bs{x}_1)}
    \left[ \|{\bs{v}_t(\bs{x}, t) - \bs{u}_t(\bs{x} \mid \bs{x}_1)}\|^{2}_{G}\right
    ],
\end{equation}
where $\| \cdot \|^{2}_{G}$ is the norm induced by the Riemannian metric $G$.
Then the learned vector field $\bs{v}_{t}$ can be used to generate samples on
the manifold at inference, which is $\mathrm{SE}(3)$ poses of the fragments. The
rigid motion of fragments corresponds to the geodesic on the \textit{Lie group} $\mathrm{SE}
(3)$, a differentiable Riemannian manifold.

\subsection{More attention on large fragments}
\acronym{} provides tailored designs to place more attention on
  large fragments. We observed that: (i) large fragments are easier to assemble; (ii)
  tiny fragments sometimes lead to unstable training. Driven by these insights, we
  apply weighted sampling based on the surface area of fragments and modify the self-attention
  module to allow more attention on large fragments. 
    \begin{figure}[h]
    \vspace{-10pt}
    \centering
    \includegraphics[width=1\linewidth]{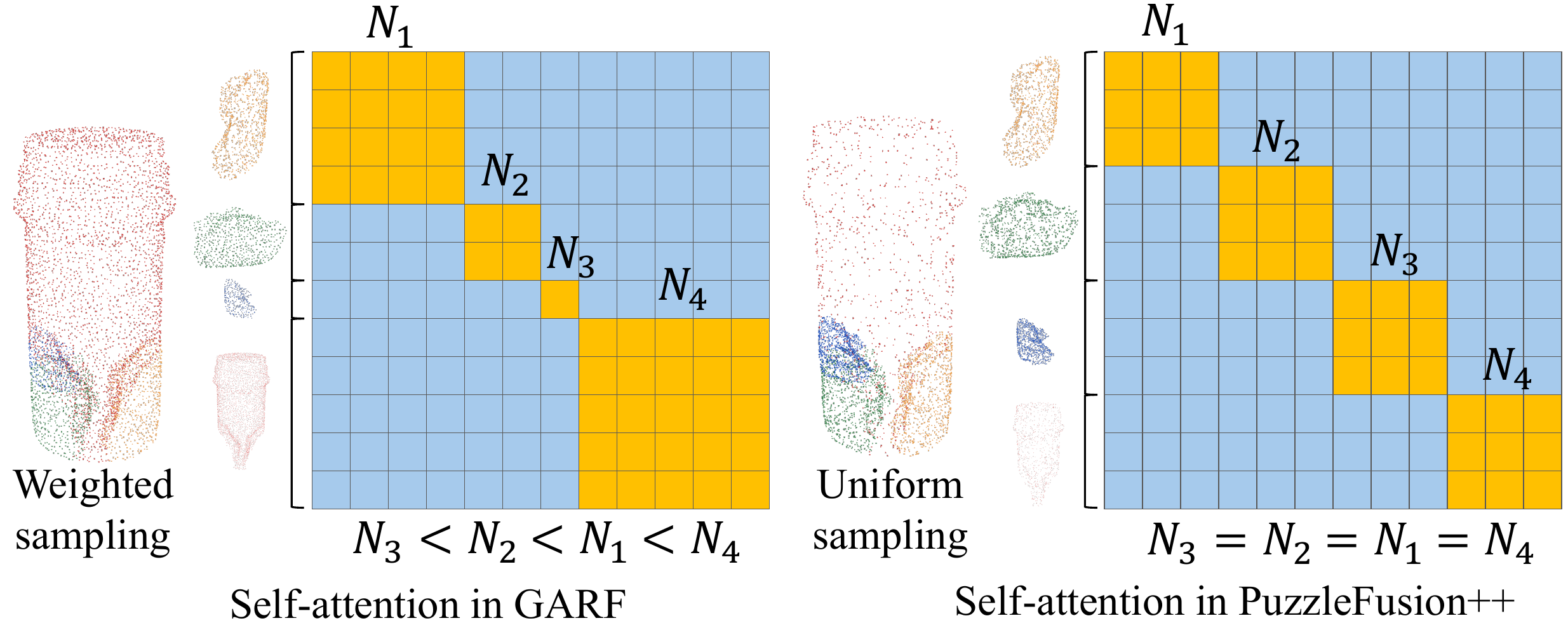}
    \caption{Self-attention comparison between \acronym{} (left) and PuzzleFusion++~\cite{wang2024puzzlefusion++} (right).}
    \vspace{-20pt}
  \end{figure}

\begin{table}[t]
    \caption{Results on Vanilla Breaking Bad~\cite{sellan2022breaking} Dataset.}
    \centering
    \scalebox{0.80}{
    \begin{tabular}{ccccc}
        \toprule \multirow{2}{*}{Methods}                                                           & RMSE(R) $\downarrow$ & RMSE(T) $\downarrow$ & PA $\uparrow$  & CD $\downarrow$  \\
                                                                                                    & \text{degree}        & $\times 10^{-2}$     & \%             & $\times 10^{-3}$ \\
        \midrule \multicolumn{5}{>{\columncolor{green!10}}c}{Tested on the \textbf{Everyday} Subset} \\
        \midrule Global~\cite{li2020learning}                                                       & 80.70                & 15.10                & 24.60          & 14.60            \\
        LSTM~\cite{wu2020pq}                                                                        & 84.20                & 16.20                & 22.70          & 15.80            \\
        DGL~\cite{zhan2020generative}                                                               & 79.40                & 15.00                & 31.00          & 14.30            \\
        SE(3)-Equiv~\cite{wu2023leveraging}                                                         & 79.30                & 16.90                & 8.41           & 28.50            \\
        DiffAssemble~\cite{scarpellini2024diffassemble}                                             & 73.30                & 14.80                & 27.50          & -                \\
        PHFormer~\cite{cui2024phformer}                                                             & 26.10                & 9.30                 & 50.70          & 9.60             \\
        Jigsaw~\cite{lu2024jigsaw}                                                                  & 42.30                & 10.70                & 57.30          & 13.30            \\
        PF++~\cite{wang2024puzzlefusion++}                                                          & 38.10                & 8.04                 & 70.60          & 6.03             \\
        \textbf{\acronym{}-mini }                                                                   & \textbf{10.41}       & \textbf{1.91}        & \textbf{92.77} & \textbf{0.45}    \\
        \midrule \multicolumn{5}{>{\columncolor{green!10}}c}{Tested on the \textbf{Artifact} Subset} \\
        \midrule Jigsaw                                                                             & 52.40                & 22.20                & 45.60          & 14.30            \\
        PF++                                                                                        & 52.10                & 13.90                & 49.60          & 14.50            \\
        \textbf{\acronym{}-mini }                                                                   & \textbf{11.91}       & \textbf{2.74}        & \textbf{89.42} & \textbf{1.05}    \\
        \bottomrule
    \end{tabular}
    } \label{table:bbv_results}
\end{table}

\begin{table}[t]
    \caption{\textbf{Ablation Study on Our Designs of FM.} }
    \label{table:fmdesigns}
    \vspace{-5pt}
    \centering
    \scalebox{0.73}{
    \begin{tabular}{cccccc}
        \toprule $\mathrm{SE}(3)$ & Multi-Anchor   & One-Step       & RMSE(R) $\downarrow$ & RMSE(T) $\downarrow$ & PA $\uparrow$  \\
        \midrule                   
        $\usym{2613}$             & $\usym{2613}$  & $\usym{2613}$  & 10.24                & 1.95                 & 89.08          \\
        $\usym{1F5F8}$            & $\usym{2613}$  & $\usym{2613}$  & 8.02                 & 1.63                 & 93.78          \\
        $\usym{1F5F8}$            & $\usym{1F5F8}$ & $\usym{2613}$  & 7.63                 & 1.60                 & 94.02          \\
        $\usym{1F5F8}$            & $\usym{1F5F8}$ & $\usym{1F5F8}$ & \textbf{6.68}        & \textbf{1.34}        & \textbf{94.77} \\
        \bottomrule
    \end{tabular}
    }
\end{table}

\begin{table}[t]
    \caption{\textbf{Ablation Study on Sample Steps.} }
    \label{table:samplesteps}
    \vspace{-5pt}
    \centering
    \scalebox{0.72}{
    \begin{tabular}{ccccccc}
        \toprule Steps         & RMSE(R) $\downarrow$ & RMSE(T) $\downarrow$ & PA $\uparrow$  & CD $\downarrow$ & Speed (ms) \\
        \midrule 1             & 12.52                & 3.18                 & 86.88          & 2.14            & 38.26      \\
        One-Step + 1           & 9.79                 & 2.46                 & 91.31          & 1.42            & 45.76      \\
        5                      & 8.25                 & 1.92                 & 93.70          & 0.53            & 57.32      \\
        One-Step + 5           & 7.15                 & 1.66                 & 94.43          & 0.46            & 76.23      \\
        20                     & 7.63                 & 1.60                 & 94.02          & 0.35            & 185.05     \\
        \textbf{One-step + 20} & \textbf{6.68}        & \textbf{1.34}        & \textbf{94.77} & \textbf{0.25}   & 190.77     \\
        50                     & 7.50                 & 1.54                 & 94.01          & 0.32            & 408.40     \\
        \bottomrule
    \end{tabular}
    }
\end{table}

\begin{table}[t]
    \caption{\textbf{Ablation Study on the Different Anchor Initialization.} }
    \label{table:anchor}
    \vspace{-5pt}
    \centering
    \scalebox{0.75}{
    \begin{tabular}{cccccc}
        \toprule Settings       & RMSE(R) $\downarrow$ & RMSE(T) $\downarrow$ & PA $\uparrow$ & CD $\downarrow$ \\
        \midrule Largest Anchor & 6.10                 & 1.22                 & 95.33         & 0.22            \\
        Random Anchor           & 6.09                 & 1.30                 & 95.20         & 0.29            \\
        Anchor-Free             & 9.09                 & 2.13                 & 93.23         & 0.91            \\
        \bottomrule
    \end{tabular}
    }
\end{table}

\begin{table}[t]
    \caption{\textbf{Comparison Between Diffusion and Our FM Models.} }
    \label{table:comparison}
    \vspace{-5pt}
    \centering
    \scalebox{0.75}{
    \begin{tabular}{cccccc}
        \toprule Dataset                     & Methods                    & RMSE(R) $\downarrow$ & RMSE(T) $\downarrow$ & PA $\uparrow$    \\
        \midrule \multirow{5}{*}{Everyday}   & Diffusion                  & 7.45                 & 1.47                 & 94.30            \\
                                             & $\mathrm{SE}(3)$ Diffusion & N/A                  & N/A                  & N/A              \\
                                             & Diffusion w/ One-Step      & 7.51                 & 1.47                 & 94.27            \\
                                             & Vanilla FM                 & 10.24                & 1.95                 & 89.08            \\
                                             & \textbf{\acronym{}-mini}   & \textbf{6.68}        & \textbf{1.34}        & \textbf{94.77}   \\
        \midrule \multirow{2}{*}{\dataset{}} & Diffusion                  & $32.38$              & $7.90$               & $71.73$          \\
                                             & \textbf{\acronym{}-mini}   & $\textbf{27.88}$     & $\textbf{6.79}$      & $\textbf{76.25}$ \\
        \bottomrule
    \end{tabular}
    }
    \vspace{-10pt}
\end{table}

\section{Additional Results and Analyses}
\label{sec:exp}
\subsection{Ablation on Design Choices in Flow Matching}
We conduct an ablation study to evaluate the impact of design choices in our FM module.
As shown in Table~\ref{table:fmdesigns}, vanilla FM, trained with spherical
linear interpolation (slerp) to approximate valid rotations in the forward
process~\cite{guo2025assembleflow}, achieves 89.08 PA, already surpassing
previous methods~\cite{wang2024puzzlefusion++, lu2024jigsaw}. Incorporating the $\mathrm{SE}
(3)$ representation further improves performance by pre-modeling the manifold distribution
and better capturing distribution shifts during assembly. Multi-anchor training
strategy further enhances results, while one-step pre-assembly significantly boosts
performance by providing a more reasonable initial pose distribution, leading to
the best overall outcomes.

\subsection{Ablation on Sample Steps}
Table \ref{table:samplesteps} shows the effect of varying sampling steps in our
framework. Surprisingly, even with just 5 steps, FM achieves 93.70\% PA, highlighting
its effectiveness in modeling global probabilistic paths. Additionally, our first-session
initialization provides a more reasonable initial pose, further improving
assembly quality while adding minimal computational overhead.

\subsection{Ablation on Anchor Fragment}
Similar to PF++~\cite{wang2024puzzlefusion++}, we use the largest fragment as
the anchor fragment at inference. We compare the performance of using the largest
fragment, a randomly selected fragment, and no anchor fragment. As shown in
Table~\ref{table:anchor}, using a random fragment as the anchor fragment has almost
no negative effect on the model. Only anchor-free initialization leads to a
slight performance drop.

\subsection{Comparison with Diffusion Models}
Table \ref{table:comparison} compares our FM module with diffusion models. While
diffusion, when paired with fracture-aware pretraining, achieves competitive performance,
directly applying vanilla FM yields lower results (89.08 PA), emphasizing the importance
of our subsequent design choices. A key limitation of diffusion models is their
handling of $\mathrm{SO}(3 )$ rotation, which cannot be naturally incorporated into
the reverse process. Existing methods, such as score prediction~\cite{yim2023se},
aim to maintain rotation validity but fall outside our current scope. Additionally,
diffusion models rely on multi-step denoising without explicitly modeling the
global probabilistic path, rendering one-step pre-assembly ineffective.
Furthermore, on \dataset{}, diffusion models exhibit weaker generalization to unseen
objects compared to \acronym{}-mini. 

\subsection{Quantitative Results on Vanilla Breaking Bad}
Given that all our previous experiments were conducted on the volume-constrained
version of the Breaking Bad dataset~\cite{sellan2022breaking}, we here provide
additional quantitative results on the non-volume-constrained version to align
with the settings of previous methods. The results, shown in Table~\ref{table:bbv_results},
demonstrate that our \acronym{}-mini model still significantly outperforms the
previous state-of-the-art method, PF++~\cite{wang2024puzzlefusion++}, by a large
margin. This performance is consistent across both the everyday and artifact subsets,
showcasing the model's robust generalization ability.

We also present the results of FragmentDiff~\cite{xu2024fragmentdiff} on their
custom Breaking Bad dataset in Table~\ref{table:fragmentdiff_results}. FragmentDiff
claims to remove tiny pieces, but it is unclear whether this applies only to
their training setting or also to evaluation. Unfortunately, since they did not
open source their code or provide their preprocessed data, we are unable to directly
compare all other methods with FragmentDiff. Additionally, they did not adhere to
the common settings used by other methods, which limit the number of pieces from
2 to 20, making direct comparisons on their provided metrics impossible. However,
its significant performance drop from the Everyday subset to the Artifact subset
suggests that \acronym{} surpasses FragmentDiff in generalization capability.

\subsection{Quantitative Results of Finetuning on the \textbf{\dataset{}} Synthetic
Dataset}
After finetuning \acronym{} on the \dataset{} synthetic dataset, we report the per-category
performance on the bone and eggshell categories, as shown in Table~\ref{table:finetuning_results}.
The results demonstrate that finetuning the FM model in \acronym{} significantly
improves performance on these two unseen categories, showing the effectiveness of
our finetuning techniques and the generalizability of our pretraining strategy.

\subsection{Additional Qualitative Comparison on the \textbf{\dataset{}} and Breaking
Bad Dataset}

Figures~\ref{fig:supp1},~\ref{fig:supp2} and~\ref{fig:supp3} demonstrate more qualitative
comparison on the \dataset{} and Breaking Bad Dataset, where our \acronym{}
shows superior performance than the other previous SOTA methods.


\begin{table}[t]
    \caption{FragmentDiff~\cite{xu2024fragmentdiff} Results on Their Custom
    Breaking Bad Dataset.}
    \centering
    \scalebox{0.80}{
    \begin{tabular}{ccccc}
        \toprule \multirow{2}{*}{Methods}                                & \multirow{2}{*}{Subset} & RMSE(R) $\downarrow$ & RMSE(T) $\downarrow$ & PA $\uparrow$ \\
                                                                         &                         & \text{degree}        & $\times 10^{-2}$     & \%            \\
        \midrule \multirow{2}{*}{FragmentDiff~\cite{xu2024fragmentdiff}} & Everyday                & 13.68                & 7.41                 & 90.20         \\
                                                                         & Artifact                & 18.18                & 8.12                 & 82.30         \\
        \bottomrule
    \end{tabular}
    } \label{table:fragmentdiff_results}
\end{table}

\vspace{-5pt}

\begin{table}[t]
    \caption{Quantitative Per-category Results on the \dataset{} (Synthetic
    Fracture).}
    \centering
    \scalebox{0.73}{
    \begin{tabular}{cccccc}
        \toprule \multirow{2}{*}{Category} & \multirow{2}{*}{Method}             & RMSE(R) $\downarrow$ & RMSE(T) $\downarrow$ & PA $\uparrow$  & CD $\downarrow$  \\
                                           &                                     & \text{degree}        & $\times 10^{-2}$     & \%             & $\times 10^{-3}$ \\
        \midrule \multirow{4}{*}{Bone}     & Jigsaw                              & 66.44                & 20.54                & 27.24          & 91.70            \\
                                           & PF++                                & 66.28                & 20.50                & 29.81          & 47.78            \\
                                           & \acronym{}                          & 17.70                & 3.80                 & 85.18          & 5.11             \\
                                           & \textbf{$\acronym{}_{\text{LoRA}}$} & \textbf{8.79}        & \textbf{1.10}        & \textbf{98.19} & \textbf{0.34}    \\
        \midrule \multirow{4}{*}{Eggshell} & Jigsaw                              & 44.44                & 12.88                & 49.03          & 10.49            \\
                                           & PF++                                & 54.81                & 13.81                & 61.36          & 1.50             \\
                                           & \acronym{}                          & 22.48                & 6.16                 & 83.41          & 0.67             \\
                                           & \textbf{$\acronym{}_{\text{LoRA}}$} & \textbf{7.10}        & \textbf{1.95}        & \textbf{95.68} & \textbf{0.26}    \\
        \bottomrule
    \end{tabular}
    } \label{table:finetuning_results}
\end{table}

\begin{figure*}
    \centering
    \includegraphics[width=1\linewidth]{
        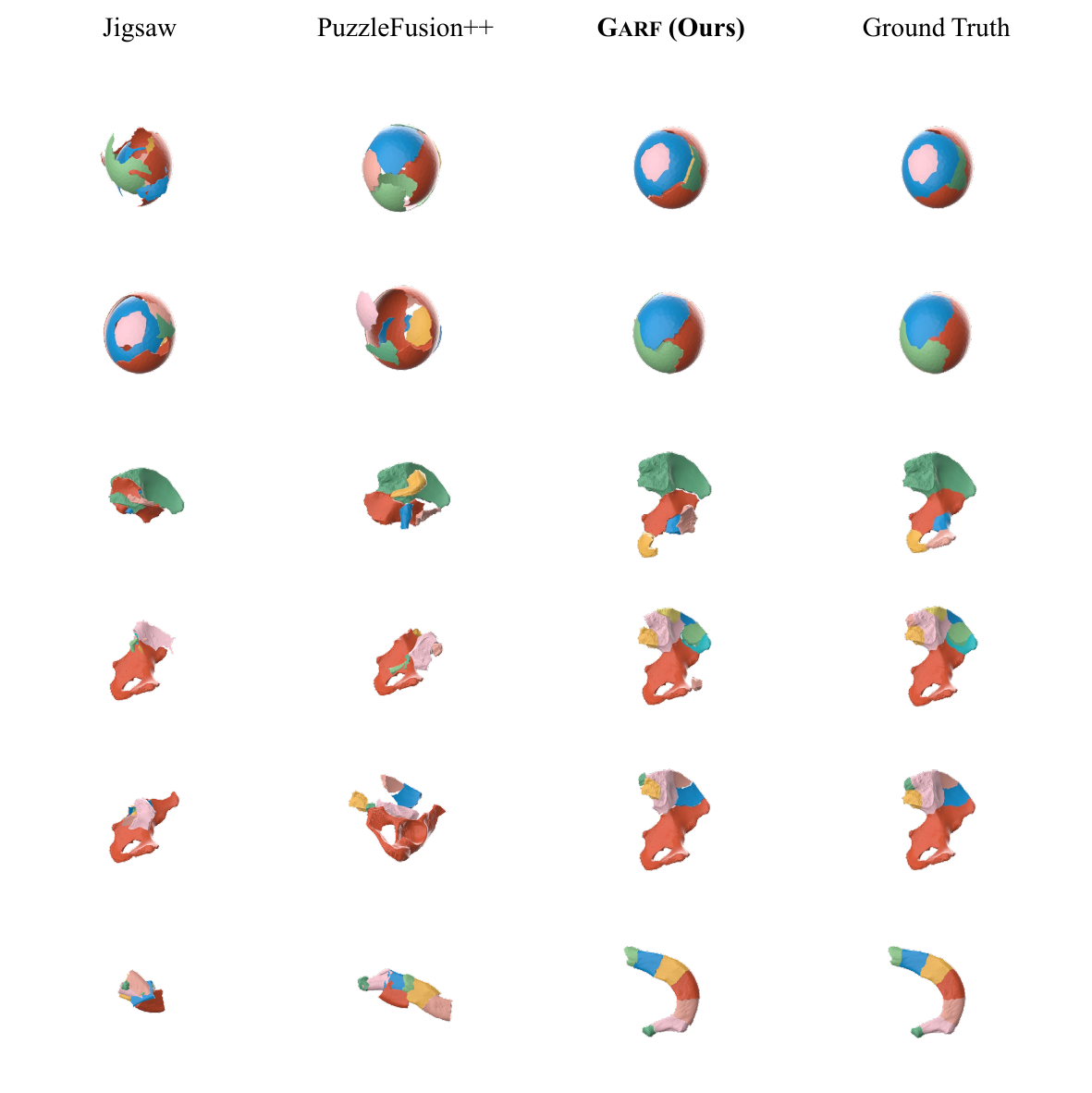
    }
    \caption{Qualitative Results on the \dataset{} Synthetic Dataset.}
    \label{fig:supp1}
\end{figure*}

\begin{figure*}
    \centering
    \includegraphics[width=1\linewidth]{
        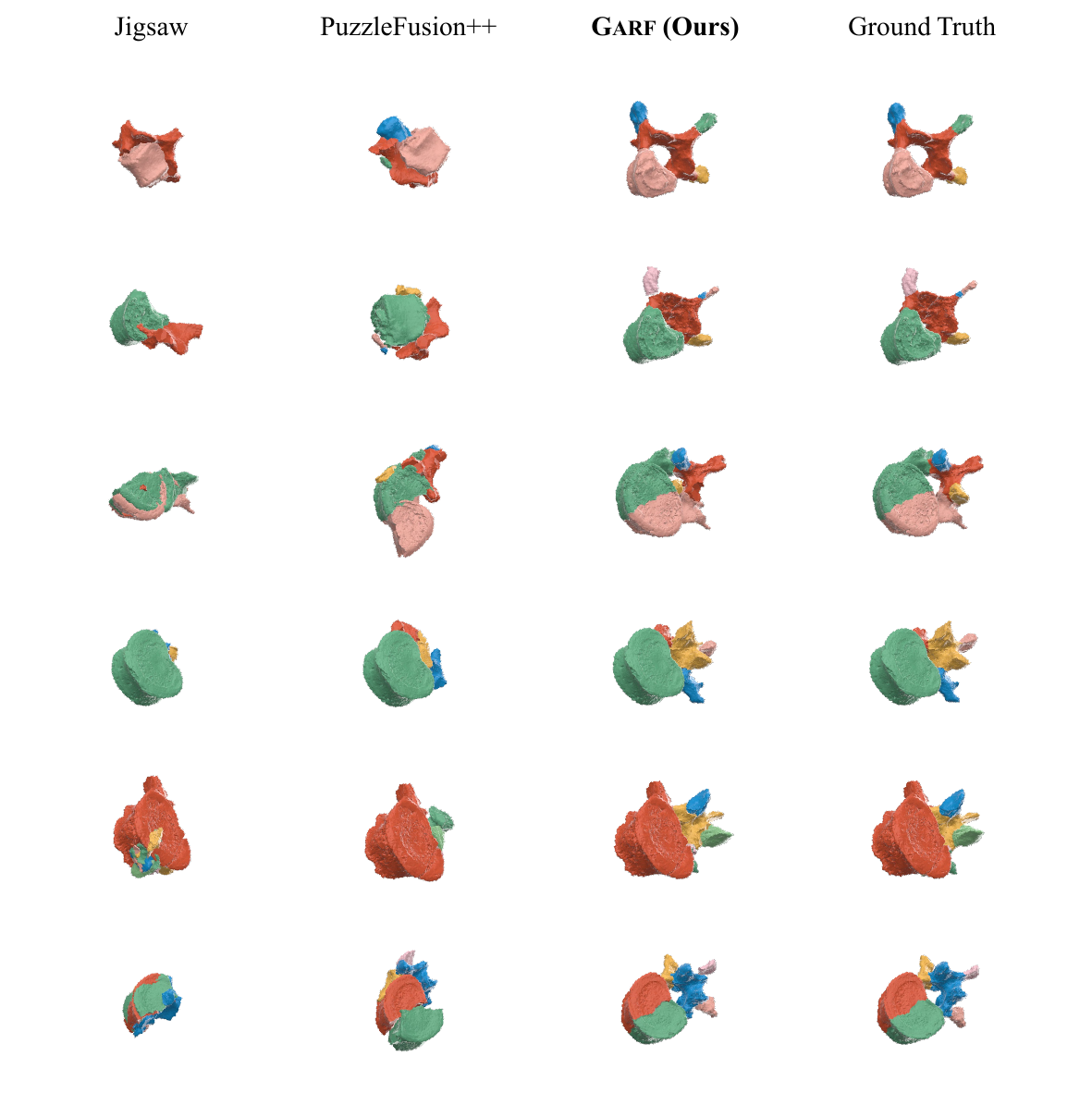
    }
    \caption{Qualitative Results on the \dataset{} Synthetic Dataset.}
    \label{fig:supp2}
\end{figure*}

\begin{figure*}
    \centering
    \includegraphics[width=1\linewidth]{
        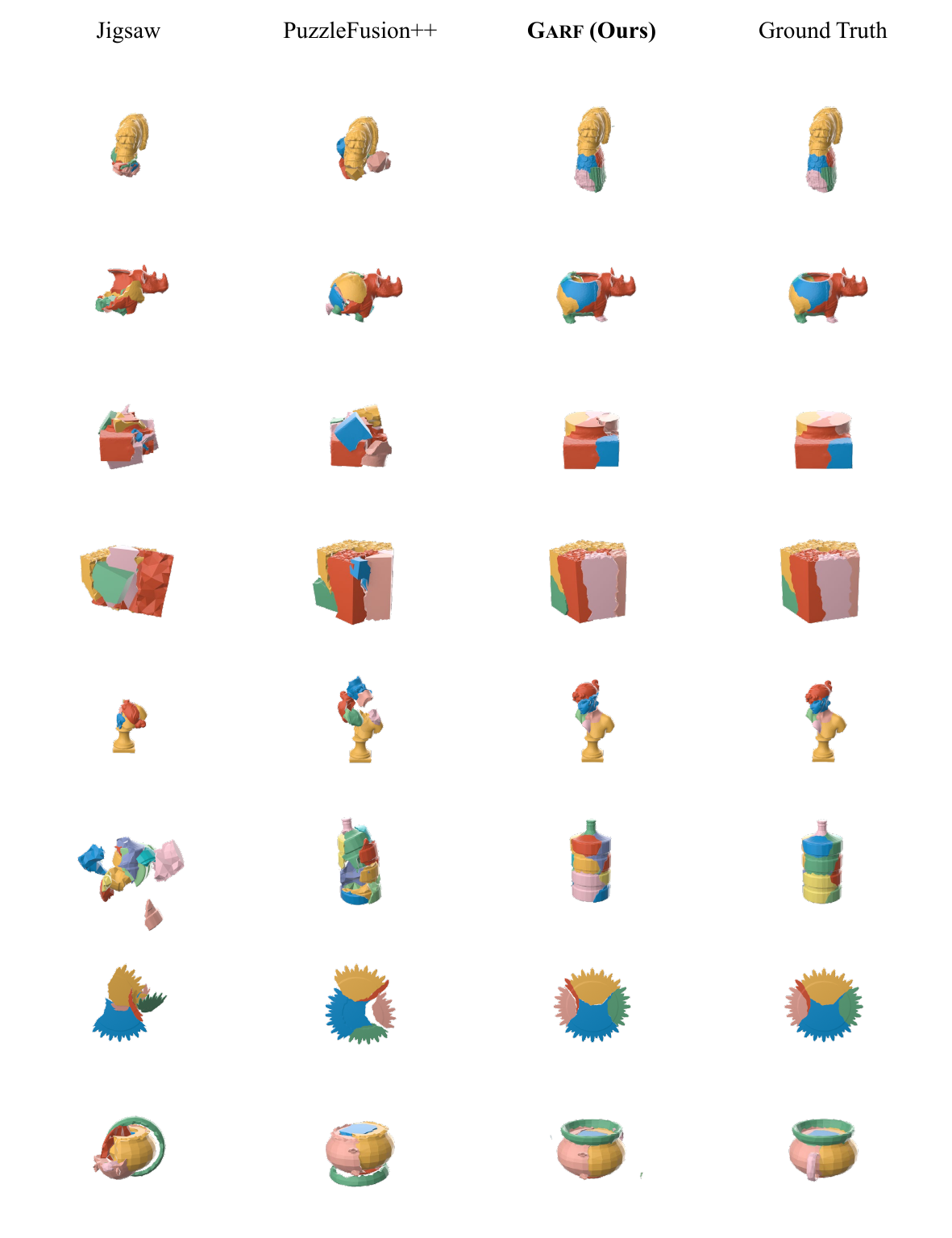
    }
    \vspace{-30pt}
    \caption{Qualitative Results on the Breaking Bad Dataset Artifact Subset.}
    \label{fig:supp3}
\end{figure*}

\end{document}